\RequirePackage{fix-cm}
\documentclass{article} 
\usepackage{iclr2027_conference,times}

\usepackage{amsmath,amsfonts,bm}

\def\eqref#1{equation~\ref{#1}}

\def\1{\bm{1}}

\DeclareMathAlphabet{\mathsfit}{\encodingdefault}{\sfdefault}{m}{sl}
\SetMathAlphabet{\mathsfit}{bold}{\encodingdefault}{\sfdefault}{bx}{n}

\usepackage{hyperref}
\usepackage{url}
\usepackage{wrapfig}

\usepackage{graphicx}
\usepackage{booktabs}
\usepackage{xcolor}
\usepackage{colortbl}
\usepackage{longtable}
\usepackage{arydshln}
\usepackage{pifont}
\usepackage{tikz}
\usepackage{amsmath}
\usetikzlibrary{arrows.meta}

\definecolor{SAGEBlue}{HTML}{1F4E79}
\definecolor{SAGEPurple}{HTML}{54174F}
\definecolor{SAGEGreen}{HTML}{154B1B}
\definecolor{SAGEGray}{HTML}{333333}

\newcommand{\SAGEMaxSpeedup}{2.55}
\newcommand{\SAGEMainSpeedupRange}{2.31--2.55}
\newcommand{\SAGEFirstTokenReductionRange}{85.5--91.9}

\title{SAGE: Pushing 2-Bit Vector-Quantized LLMs toward Real-World Usability}

\author{
  Xianglong Yan$^{1}$\thanks{Equal contribution.},\enspace
  Zenan Wu$^{1}$\footnotemark[1],\enspace
  Kewei Zhang$^{1}$,\enspace
  Kaicheng Yang$^{1}$,\enspace
  Yulun Zhang$^{1}$\thanks{Corresponding author: Yulun Zhang, \texttt{yulun100@gmail.com}.}\\
  $^{1}$Shanghai Jiao Tong University
}

\iclrfinalcopy 
\begin{document}

\maketitle

\fancyhead{}
\renewcommand{\headrulewidth}{0pt}

\vspace{-7mm}
\begin{abstract}
\vspace{-3mm}
Extreme low-bit weight-only quantization is an effective approach to reducing the deployment costs of large language models.
Below 3 bits, scalar quantization often suffers severe accuracy loss, while vector quantization (VQ) better preserves accuracy through more expressive joint encoding.
However, existing VQ methods still exhibit substantial degradation on real-world tasks, and two design challenges remain underexplored: (1) protecting sensitive weight channels within jointly encoded vectors; and (2) allocating a fixed bit budget among representation components, such as direction and radius, to suit different weight matrices.
To address these challenges, we propose \textbf{SAGE}: \textbf{S}plit \textbf{A}daptation and \textbf{G}rouping by \textbf{E}nergy for weight-only vector quantization.
First, Energy-Stratified Grouping (ESG) couples channel grouping with activation-weighted fitting, grouping channels from different energy tiers within each subvector to prioritize sensitive channels.
Second, Adaptive Bit Split (ABS) balances direction-codebook capacity and radius precision per matrix, selecting the split with the lowest activation-weighted reconstruction error under a fixed budget.
Across benchmarks covering knowledge, mathematics, code generation, instruction following, and tool use, SAGE achieves state-of-the-art performance in the 2-bit quantization setting.
On Qwen3-30B-A3B, it retains 95.3\% of the BF16 model's average benchmark score.
With our custom-designed CUDA kernel, SAGE delivers up to a $\SAGEMaxSpeedup\times$ non-streaming end-to-end speedup over FP16. The code is available at \url{https://github.com/Zincwolf/sage}.
\end{abstract}

\setlength{\abovedisplayskip}{2pt}
\setlength{\belowdisplayskip}{2pt}

\vspace{-5mm}
\section{Introduction}
\vspace{-3mm}

Recent large language models (LLMs)~\citep{kimi-k3,glm5,qwen3,dsv4,ministral3} have made remarkable progress in reasoning, coding, and generation.
Those gains have come with steadily growing parameter counts.
A larger model occupies more memory and spends more compute on every token it generates, which makes it hard to run on edge devices and other resource-constrained platforms. 
Local deployment prioritizes low-latency inference with small batches.
In this setting, decoding offers little weight reuse across requests, making memory bandwidth rather than compute the primary bottleneck~\citep{powerinfer}.

Weight-only post-training quantization (PTQ)~\citep{gptq,awq,qtip,paroquant} targets exactly this bottleneck.
Lower-bit weights shrink the memory footprint and can accelerate memory-bound decoding by reducing weight traffic~\citep{squeezellm}. Scalar methods such as GPTQ~\citep{gptq} and AWQ~\citep{awq} have made 3--4-bit weights practical with little accuracy loss.
Below 3 bits the picture changes: each weight has only a handful of reconstruction values left, and accuracy degrades sharply~\citep{aqlm}.
Vector quantization (VQ)~\citep{aqlm,gptvq,llvq} instead encodes groups of weights jointly against a shared codebook, so at the same bit budget its reconstruction points follow the weight distribution far more closely.
This is what motivates the sub-3-bit line of work built on VQ and related structured coding~\citep{quip-sharp,qtip}.

Existing VQ methods differ in how the codebook is structured and how a weight is recovered from it.
QuIP\#~\citep{quip-sharp} and QTIP~\citep{qtip} use incoherence processing, requiring online Hadamard transforms on activations during inference.
QuIP\# combines table lookups with sign and shift operations, QTIP supports lookup-only, computed, and hybrid trellis codes, and LLVQ~\citep{llvq} uses structured Leech lattice decoding without an explicit codebook.
PCDVQ~\citep{pcdvq} does use an explicit direction codebook, but a substantially
larger one.
SPHQuant~\citep{sphquant} reaches comparable accuracy and decodes considerably faster.
It stores each group of eight weights, a subvector, in sign-separated form: the signs explicitly, the magnitude through a quantized radius, and the direction as an index into a learned codebook over the non-negative orthant.
Folding all $2^8$ orthants into one keeps that codebook at 64 entries, or 1\,KiB, so it stays in shared memory and decoding is a lookup, a sign flip, and a multiply.

\begin{figure}[t]
    \centering
    \includegraphics[width=\textwidth]{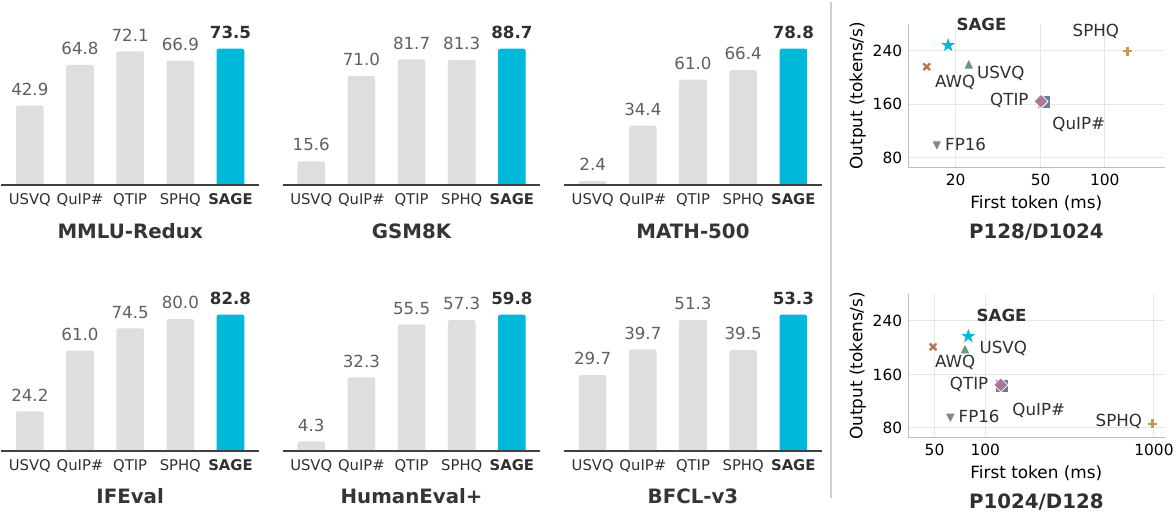}
    \vspace{-5mm}
    \caption{\textbf{Accuracy and inference efficiency on Qwen3-4B-Instruct-2507.}
    Left: generative scores, with independent zero-based axes.
    Right: batch-one non-streaming first-token latency versus output throughput on an RTX 4090; upper left is better.
    Throughput includes prompt processing. P/D denote input/output token counts. Quantized weights are reconstructed on demand.}
    \label{fig:teaser}
    \vspace{-5mm}
\end{figure}

However, two challenges remain insufficiently addressed in vector quantization. \textbf{(1) How to protect important channels.}
Protecting the channels that matter most is standard in scalar quantization, by rescaling them~\citep{awq} or by permuting them so that a group's channels are alike and share a scale well~\citep{pt2llm}.
Neither strategy transfers directly to vector quantization, where eight channels share one codeword and one radius: both rely on properties of a per-weight scalar grid that a jointly coded subvector does not have. \textbf{(2) How to divide the fixed budget.}
A group of eight weights costs 18 bits: eight signs, six for the direction index, four for the radius, a nominal 2.25 bits per weight.
How the remaining ten are split between the index and the radius is itself a choice, since moving one bit from the radius to the index doubles the codebook at almost no extra cost. Previous work adopts a fixed split~\citep{sphquant}.

Beyond quantization algorithm design, evaluating the practical capabilities of quantized models is equally important.
Common metrics such as perplexity and multiple-choice accuracy on benchmarks like HellaSwag~\citep{hellaswag} do not fully characterize the generative capabilities of post-trained LLMs.
In particular, a small increase in perplexity does not necessarily imply that a model can still reason correctly, follow instructions, or produce functional code.
We therefore evaluate quantized models on generative benchmarks covering knowledge, mathematical reasoning, code generation, instruction following, and tool use.
Our goal is to preserve the ability of extremely low-bit models to complete real-world tasks, which requires assessing these capabilities directly.

In this paper, we propose \textbf{SAGE}: \textbf{S}plit \textbf{A}daptation and \textbf{G}rouping by \textbf{E}nergy for weight-only vector quantization.
To address Challenge (1), we introduce Energy-Stratified Grouping (ESG), which couples channel grouping with activation-weighted reconstruction.
ESG distributes energy-ranked channels across subvectors, placing channels from different energy tiers within each subvector so that weighted fitting can prioritize sensitive channels.
To address Challenge (2), we introduce Adaptive Bit Split (ABS), which evaluates candidate allocations between the direction index and radius using activation-weighted reconstruction error and selects a split for each matrix.
Together, ESG and ABS adapt the representation to channel importance and matrix-specific demands while preserving the encoding budget.
Fig.~\ref{fig:teaser} illustrates both accuracy and inference efficiency on Qwen3-4B-Instruct-2507.
SAGE achieves the highest quantized scores across all six generative benchmarks and combines low first-token latency with high output throughput under both input/output length settings.
Our custom-designed CUDA kernels further enable up to a $\SAGEMaxSpeedup\times$ non-streaming end-to-end speedup over FP16. We hope SAGE advances low-bit models toward real-world deployment.

Our key contributions can be summarized as follows:
\vspace{-2mm}
\begin{itemize}
    \item We propose \textbf{SAGE}: \textbf{S}plit \textbf{A}daptation and \textbf{G}rouping by \textbf{E}nergy, a novel weight-only VQ framework that advances 2-bit quantization toward real-world usability.
    \item We design \textbf{Energy-Stratified Grouping (ESG)}, which couples channel grouping with activation-weighted fitting.
    ESG groups channels from different energy tiers within each subvector, enabling weighted fitting to prioritize sensitive channels.
    \item We develop \textbf{Adaptive Bit Split (ABS)}, which derives the division between the direction index and the radius from activation-weighted reconstruction error and adapts it per matrix, rather than applying one fixed split across the whole model, at an unchanged bit budget.
    \item Experiments on five LLMs and six generative benchmarks establish \textbf{SAGE} as a new SOTA for 2-bit VQ, retaining 95.3\% of the BF16 average score on Qwen3-30B-A3B.
    Our CUDA kernel achieves up to a $\SAGEMaxSpeedup\times$ non-streaming end-to-end speedup over FP16.
\end{itemize}

\vspace{-2mm}
\section{Related Works}
\vspace{-2mm}

Quantization can be categorized by optimization and representation.
Post-training quantization (PTQ) uses small calibration sets for calibration or lightweight blockwise reconstruction, while quantization-aware training (QAT) typically involves large-scale end-to-end training.
Scalar quantization (SQ) encodes weights individually, whereas vector quantization (VQ) encodes them jointly.
Our method is weight-only vector PTQ, and we organize this section by representation.

\vspace{-2mm}
\paragraph{Scalar Quantization.}
Among SQ approaches, PTQ is the most widely adopted, for its low calibration cost.
At 8 bits, LLM.int8()~\citep{llmint8} and SmoothQuant~\citep{smoothquant} preserve accuracy via outlier handling and channel-wise scaling, while at 4 bits, GPTQ~\citep{gptq} and AWQ~\citep{awq} reduce quantization error through second-order compensation and activation-aware scaling.
Beyond direct quantization, rotation-based methods redistribute outliers with fixed or learned orthogonal transforms~\citep{quarot,spinquant,quip,flatquant,ostquant,paroquant}, while permutation-based methods leave channels intact and merely reorder them, gathering
channels of similar range so that a shared scale fits every member of a group~\citep{pt2llm}.
To limit error accumulation, GPTAQ~\citep{gptaq} and QEP~\citep{qep} account for errors from preceding quantized layers when reconstructing layer outputs.
Others add auxiliary representations, a sparse full-precision component~\citep{squeezellm} or shared low-rank corrections~\citep{tileq}, or optimize the quantizer itself through clipping and equivalent transformations~\citep{omniquant} and loss-sensitive non-uniform grids~\citep{leanquant}.
At even lower precision, ARB-LLM~\citep{arbllm} and PT$^2$-LLM~\citep{pt2llm} extend parameter refinement to binary and ternary quantization by alternating updates to grid parameters and discrete assignments.
However, scalar PTQ can suffer substantial accuracy loss at 3 bits and below~\citep{paretoq}.
QAT addresses this through distillation~\citep{llmqat,bitdistiller} or tailored optimization~\citep{efficientqat,paretoq}, while limited scalar-grid capacity motivates more expressive VQ representations.

\vspace{-2mm}
\paragraph{Vector Quantization.}
VQ methods differ mainly in how they group weights, how they construct codebooks, and how they optimize the resulting assignments.
GPTVQ~\citep{gptvq} extends GPTQ's second-order compensation to weight vectors spanning columns; VPTQ~\citep{vptq} instead groups weights within columns for column-wise compensation.
Additive and residual codebooks further refine reconstruction~\citep{aqlm,vptq}.
QuIP\#~\citep{quip-sharp} combines incoherence processing with lattice codebooks.
QTIP~\citep{qtip} retains incoherence processing but replaces these codebooks with trellis coding, enabling higher-dimensional quantization without exponential codebook storage.
NestQuant~\citep{nestquant} develops the lattice approach using self-similar nested lattices.
A further branch applies the classical shape--gain decomposition~\citep{gersho}, coding magnitude and direction separately: PCDVQ~\citep{pcdvq} pairs a scalar magnitude code with an $E_8$ direction codebook, LLVQ~\citep{llvq} uses the Leech lattice, and SPHQuant~\citep{sphquant} additionally separates the sign pattern so that one small codebook entry serves every sign pattern.
Optimization advances include joint refinement of parameters and assignments~\citep{pvtuning} and mixed precision guided by gradients and Hessians for VLMs~\citep{mgvq}.
UniSVQ~\citep{unisvq} and LC-QAT~\citep{lcqat} use affine lattice codebooks for scalar kernels and lookup-free QAT, respectively.

\vspace{-5mm}
\section{Method}
\label{sec:method}
\vspace{-2mm}

\begin{figure}[t]
  \centering
  \includegraphics[width=\linewidth]{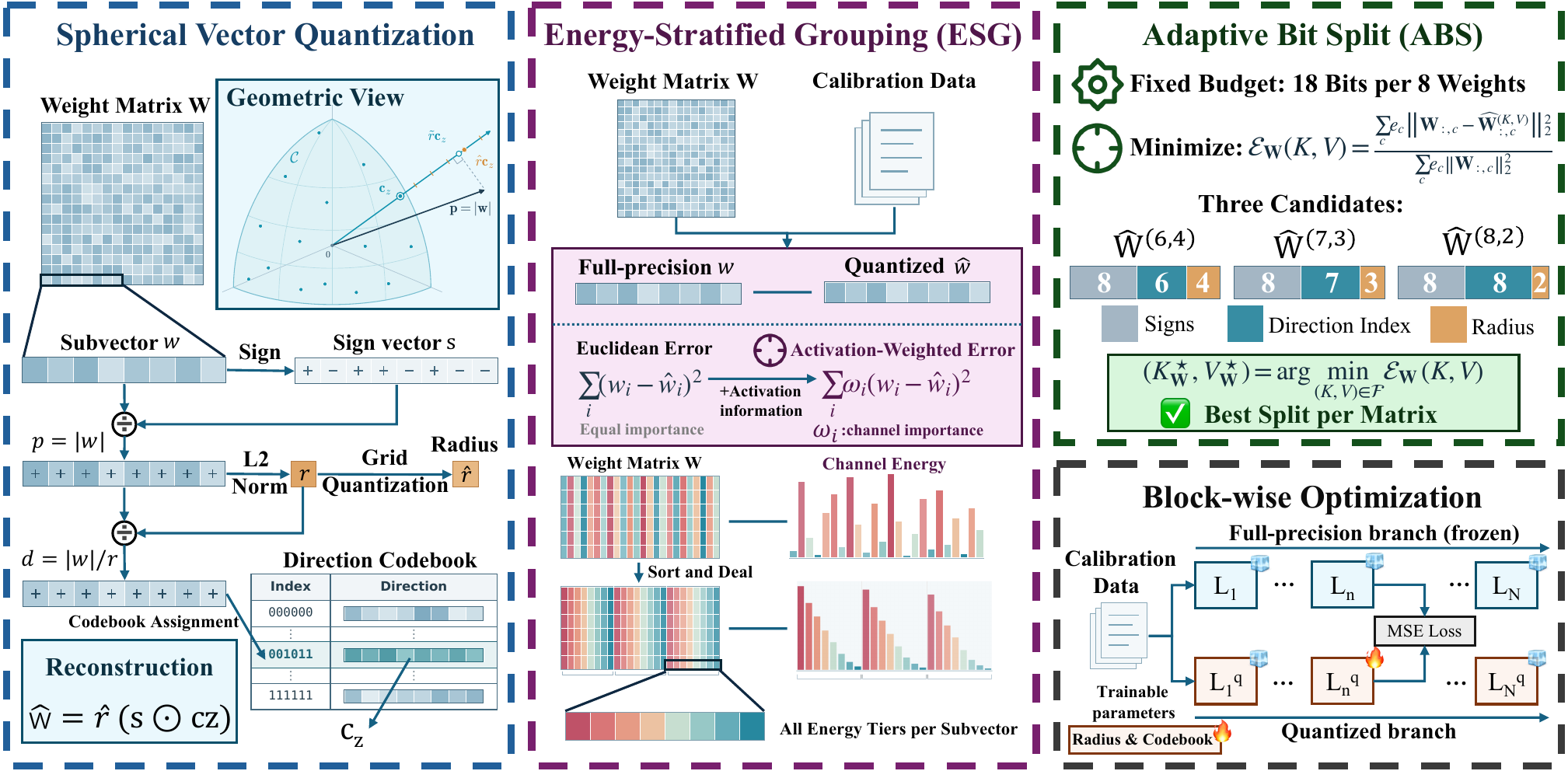}
  \vspace{-6mm}
    \caption{Overview of SAGE.
    \textcolor{SAGEBlue}{\textbf{Spherical Vector Quantization}} encodes each weight subvector using signs, a direction index, and a quantized radius.
    \textcolor{SAGEPurple}{\textbf{Energy-Stratified Grouping (ESG)}} uses calibration activations to group channels across energy tiers within subvectors and protect sensitive channels through weighted fitting.
    \textcolor{SAGEGreen}{\textbf{Adaptive Bit Split (ABS)}} compares activation-weighted reconstruction errors to select a direction-index/radius bit split per matrix, using 18 bits per eight weights.
    \textcolor{SAGEGray}{\textbf{Blockwise Optimization}} updates radius and codebook angles to match full-precision block outputs.}
  \label{fig:method}
  \vspace{-3mm}
\end{figure}

\paragraph{Overview.}
As illustrated in Fig.~\ref{fig:method}, SAGE combines activation-guided channel grouping with adaptive bit allocation for spherical vector quantization.
We first introduce the spherical representation and its bit budget in Sec.~\ref{sec:prelim}.
Building on this formulation, Sec.~\ref{sec:weighting} presents ESG to protect sensitive channels through energy-stratified grouping and weighted fitting.
Sec.~\ref{sec:split} then introduces ABS to adapt the direction-index and radius bit allocation to each weight matrix.
Finally, Sec.~\ref{sec:optimization} covers calibration and blockwise optimization, and Sec.~\ref{sec:kernel} presents our fused inference kernel.

\vspace{-2mm}
\subsection{Background: Spherical Vector Quantization}
\label{sec:prelim}
\vspace{-2mm}

\paragraph{Spherical Representation.}
Following SPHQuant~\citep{sphquant}, we partition each row of a weight matrix $\mathbf{W}\in\mathbb{R}^{d_{\mathrm{out}}\times d_{\mathrm{in}}}$ into subvectors of dimension $v=8$ along the input axis.
Each subvector $\mathbf{w}$ is decomposed into signs $\mathbf{s}=\operatorname{sign}(\mathbf{w})$, radius $r=\|\mathbf{w}\|_2$, and nonnegative unit direction $\mathbf{d}=|\mathbf{w}|/r$.
Signs are stored exactly, while the direction is represented by an entry $\mathbf{c}_z$ from a matrix-specific codebook $\mathcal{C}$ of $K$ nonnegative unit vectors.
The quantized reconstruction is:
\begin{equation}
    \widehat{\mathbf{w}}
    = \hat{r}\,(\mathbf{s}\odot\mathbf{c}_z),
    \label{eq:reconstruction}
\end{equation}
where $\hat{r}$ is the quantized radius and $\odot$ denotes elementwise multiplication.

\vspace{-2mm}
\paragraph{Direction and Radius Quantization.}
In this formulation, the codebook is initialized using spherical $k$-means, and directions are selected by cosine similarity, $z=\arg\max_k \mathbf{d}^{\top}\mathbf{c}_k$.
Given the selected direction, the continuous radius $\tilde{r}=|\mathbf{w}|^{\top}\mathbf{c}_z$ minimizes Euclidean reconstruction error.
Every $G=128$ weights share a uniform radius grid with offset $\mu_g$ and step $\delta_g$:
\begin{equation}
    \hat{r}=\mu_g+q\delta_g,
    \qquad
    q=\operatorname{clip}\!\left(
    \left\lfloor\frac{\tilde{r}-\mu_g}{\delta_g}\right\rceil,
    0,2^V-1\right),
    \label{eq:radiusgrid}
\end{equation}
where $V$ is the radius bit width, and the grid spans each group's fitted radius range.

\vspace{-2mm}
\paragraph{Bit Budget.}
Each subvector stores $v$ sign bits, a $\log_2 K$-bit direction index, and a $V$-bit radius code.
The total encoding budget is:
\begin{equation}
    B=v+\log_2 K+V.
    \label{eq:budget}
\end{equation}
The baseline configuration $(K,V)=(64,4)$ uses 18 bits per subvector, or 2.25 bits per weight excluding shared metadata.
At this fixed budget, direction capacity and radius precision can be traded by varying $K$ and $V$. Further details on spherical quantization are provided in Appendix~\ref{sec:prelim_appendix}. We next introduce Energy-Stratified Grouping (ESG) and Adaptive Bit Split (ABS).

\vspace{-2mm}
\subsection{Energy-Stratified Grouping}
\label{sec:weighting}
\vspace{-2mm}

\paragraph{Activation-Weighted Reconstruction.}
The Euclidean reconstruction criterion in Sec.~\ref{sec:prelim} treats all positions within a subvector equally.
Motivated by activation-aware channel protection~\citep{awq} and data-aware codebook fitting~\citep{gptvq}, we incorporate input activation statistics into the reconstruction objective to guide weight quantization.
For the linear layer input $\mathbf{x}$, we estimate channel energy as $e_c=\mathbb{E}[x_c^2]$ over calibration tokens, where $x_c$ is the activation of input channel $c$.
For channels $a_1,\dots,a_v$ in a subvector, we define the weights and objective as:
\begin{equation}
    \omega_i =
    \frac{e_{a_i}^{\gamma}}
    {\frac{1}{v}\sum_{j=1}^{v}e_{a_j}^{\gamma}},
    \qquad
    \mathcal{L}_{\boldsymbol{\omega}}(\mathbf{p},\tilde{r},\mathbf{c}_k)
    = \sum_{i=1}^{v}\omega_i(p_i-\tilde{r}c_{k,i})^2,
    \label{eq:weighted-objective}
\end{equation}
where $\mathbf{p}=|\mathbf{w}|$, $\tilde{r}$ is the continuous radius before quantization, and $c_{k,i}$ is the $i$-th component of codeword $\mathbf{c}_k$.
Weights are normalized to unit mean per subvector.
We use $\gamma=0.75$ to temper energy differences while prioritizing high-energy channels, with ablations in Fig.~\ref{fig:sensitivity}(a).

\begin{figure}[t]
    \centering
    \includegraphics[width=\textwidth]{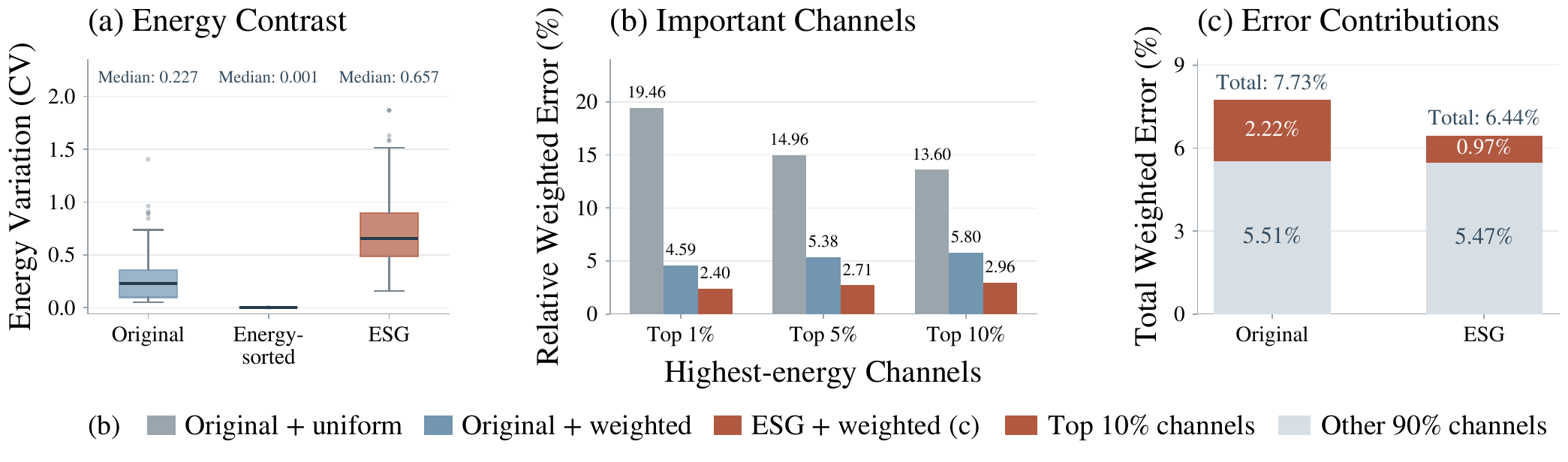}
    \vspace{-8mm}
    \caption{ESG analysis on Qwen3-4B-Instruct-2507:
    (a) within-subvector energy variation,
    (b) high-energy channel errors, and
    (c) total error contributions.}
    \label{fig:esg-analysis}
    \vspace{-5mm}
\end{figure}

\vspace{-2mm}
\paragraph{Energy-Stratified Grouping.}
The effect of weighting depends on which channels share a subvector.
When their energies are equal, $\omega_i=1$ and the objective reduces to Euclidean reconstruction.
This motivates grouping channels from different energy levels, allowing the weighted fit to prioritize important positions within each jointly encoded vector.
Let $\sigma$ sort input channels by decreasing energy, and divide the sorted list into $v$ tiers of $m=d_{\mathrm{in}}/v$ channels.
ESG constructs each subvector by taking one channel from each tier, giving the permutation:
\begin{equation}
    \pi\bigl((j-1)v+t\bigr)
    = \sigma\bigl((t-1)m+j\bigr),
    \qquad
    j=1,\dots,m,\quad t=1,\dots,v.
    \label{eq:dealing}
\end{equation}
Each subvector therefore contains channels spanning the energy ranking, with positions consistently ordered from higher to lower energy.
Fig.~\ref{fig:esg-analysis}(a) shows that ESG increases within-subvector energy variation compared with both the original order and grouping adjacent energy-ranked channels.
We apply $\pi$ to the columns of $\mathbf{W}$ and reorder the activation statistics accordingly before fitting the quantizer.
This preserves each channel's identity while changing which weights share a direction and radius.
At inference, the corresponding permutation is applied to the input activations.

\vspace{-2mm}
\paragraph{Weighted Quantizer Fitting.}
We fit the codebook using weighted spherical $k$-means in the permuted channel order.
For a candidate codeword $\mathbf{c}_k$, minimizing Eq.~(\ref{eq:weighted-objective}) over the continuous radius gives a closed-form solution.
Substituting it into the objective yields the direction assignment:
\begin{equation}
    \tilde{r}_k^{\star}
    = \frac{\sum_i \omega_i p_i c_{k,i}}
           {\sum_i \omega_i c_{k,i}^{2}},
    \qquad
    z = \arg\max_k
    \frac{\left(\sum_i \omega_i p_i c_{k,i}\right)^2}
         {\sum_i \omega_i c_{k,i}^{2}}.
    \label{eq:weighted-assign}
\end{equation}
The selected radius $\tilde{r}_z^{\star}$ is quantized using the shared grid in Eq.~(\ref{eq:radiusgrid}).
With uniform weights, the assignment reduces to the cosine rule in Sec.~\ref{sec:prelim}.
At initialization with $(K,V)=(64,4)$, ESG reduces high-energy channel errors compared with weighted fitting without reordering (Fig.~\ref{fig:esg-analysis}(b)).
Most of the overall weighted-error reduction comes from the highest-energy 10\% of channels (Fig.~\ref{fig:esg-analysis}(c)), supporting the role of grouping in protecting sensitive channels.

\vspace{-2mm}
\subsection{Adaptive Bit Split}
\label{sec:split}
\vspace{-2mm}

ESG changes the structure of weight subvectors and their reconstruction priorities, potentially altering the balance between direction and radius errors.
Under a fixed encoding budget, increasing direction-codebook capacity requires reducing radius precision.
Since the benefit of this trade-off depends on each matrix's weight distribution and activation statistics, Adaptive Bit Split (ABS) selects a matrix-specific split by comparing activation-weighted reconstruction errors at initialization.

\paragraph{Candidate Bit Splits.}
With $v=8$ and $B=18$, we consider three splits of the remaining ten bits:
\begin{equation}
    \mathcal{F}=\{(64,4),(128,3),(256,2)\},
    \qquad \log_2 K+V=10.
    \label{eq:split-candidates}
\end{equation}
All candidates use the same 2.25-bit encoding budget per weight, with a fixed radius group size and grid-parameter precision.
For each candidate, we fit the codebook and assign directions and continuous radii in the ESG-permuted order, following Sec.~\ref{sec:weighting}.
Radii are quantized using Eq.~(\ref{eq:radiusgrid}), accounting for both direction approximation and radius discretization.

\vspace{-2mm}
\paragraph{Activation-Aware Selection.}
Let $\widehat{\mathbf{W}}^{(K,V)}$ denote the reconstruction in original channel order, and $\mathbf{W}_{:,c}$ the weight column for input channel $c$.
Using calibration energies $e_c$, we compute:
\begin{equation}
    \mathcal{E}_{\mathbf{W}}(K,V)
    =
    \frac{
        \sum_c e_c
        \left\|\mathbf{W}_{:,c}
        -\widehat{\mathbf{W}}^{(K,V)}_{:,c}\right\|_2^2
    }{
        \sum_c e_c \left\|\mathbf{W}_{:,c}\right\|_2^2
    }.
    \label{eq:split-score}
\end{equation}
The numerator approximates the expected squared output error by ignoring correlations between input channels.
Fitting uses tempered, normalized weights $\boldsymbol{\omega}$, while scoring uses raw energies $e_c$ to preserve relative importance across subvectors.
The selected split is:
\begin{equation}
    (K_{\mathbf{W}}^\star,V_{\mathbf{W}}^\star)
    =
    \arg\min_{(K,V)\in\mathcal{F}}
    \mathcal{E}_{\mathbf{W}}(K,V).
    \label{eq:split-selection}
\end{equation}
Equal candidate payloads allow independent selection for each matrix while preserving the total encoding budget.
The split is fixed after initialization throughout blockwise optimization.

\vspace{-2mm}
\subsection{Calibration Data and Blockwise Optimization}
\label{sec:optimization}
\vspace{-2mm}

\paragraph{Calibration Data Construction.}
Since ESG and ABS rely on calibration activations, the calibration inputs should reflect the contexts encountered during deployment, including both user prompts and model-generated responses.
Following the use of model-generated data in LLM-QAT~\citep{llmqat} and target-model reasoning traces in AYOT~\citep{ayot}, we construct calibration sequences from the full-precision model's own responses.
We sample prompts spanning mathematics, code, reasoning, instruction following, tool use, and general conversation, apply the target model's chat template, and generate responses using greedy decoding.
Each prompt is concatenated with its response, and the resulting token streams are packed into fixed-length sequences for activation statistics collection and blockwise optimization. Further details are provided in Appendix~\ref{appendix_calibration_data}.

\vspace{-2mm}
\paragraph{Blockwise Optimization.}
Following SPHQuant, we optimize blocks sequentially against full-precision outputs, with student inputs from preceding quantized blocks.
For block $\ell$, we minimize:
\begin{equation}
    \mathcal{L}_{\mathrm{block}}
    =
    \mathbb{E}_{\mathcal{D}_{\mathrm{cal}}}
    \left[
        \left\|
        \widehat{F}_{\ell}(\mathbf{x}_{\ell}^{\mathrm{q}})
        -
        F_{\ell}(\mathbf{x}_{\ell}^{\mathrm{fp}})
        \right\|_2^2
    \right],
    \label{eq:blockwise-loss}
\end{equation}
where $F_{\ell}$ and $\widehat{F}_{\ell}$ denote the full-precision and quantized blocks, respectively.
We jointly update the continuous radius parameters and codebook angles while keeping channel permutations, bit splits, and direction indices fixed.
Radius quantization uses a straight-through estimator, and angular parameterization preserves nonnegative unit codewords.
This procedure allows each block to compensate for errors introduced by preceding quantized blocks.

\vspace{-2mm}
\subsection{CUDA Kernel Design for Efficient Inference}
\label{sec:kernel}
\vspace{-2mm}

\paragraph{Channel Permutation and Fusion.}
ESG stores each weight matrix in permuted input-channel order; each projection applies its corresponding permutation to the input activations.
In compiled decoding, we keep permutation visible to the graph compiler for fusion with preceding normalization or activation operations.
This avoids repeating the gather in every output thread block.
Eager decoding fuses gather into matrix--vector multiplication for short inputs, avoiding a separate launch.
Prefill reorders activations once per projection for reuse across output tiles.

\vspace{-2mm}
\paragraph{Quantized Matrix Multiplication with Adaptive Bit Splits.}
All ABS splits share a 16-bit main word plus two auxiliary bits per subvector, with compile-time specialization of bit extraction.
During decoding, contiguous shared-memory codewords enable 128-bit loads for register-local packed FP16 dot products, followed by FP32 radius weighting and reduction.
Shape-dependent output tiling balances activation reuse against register pressure and thread-block parallelism.
During prefill, Tensor Cores reuse weight tiles decoded into shared memory across 32 or 64 activation rows.
The larger tile amortizes decoding, while staging only 64 activation channels at a time limits shared-memory usage.
For small grids, we expose more parallelism by splitting the reduction dimension across thread blocks, then combine partial sums in FP32. Further details are provided in Appendix~\ref{appendix_kernel_details}.

\vspace{-2mm}
\section{Experiments}
\vspace{-2mm}
\label{sec:exp}

\begin{table}[!t]
\centering
\caption{Weight-only vector quantization results on six benchmarks (higher is better).
\textbf{Bold} indicates the best quantized result for each model and metric.
BF16 denotes the full-precision model.}
\label{tab:main-results}
\small
\setlength{\tabcolsep}{4pt}
\renewcommand{\arraystretch}{1.08}
\setlength{\dashlinedash}{2pt}
\setlength{\dashlinegap}{2pt}
\resizebox{\textwidth}{!}{%
\begin{tabular}{l*{8}{c}}
\toprule
\rowcolor{blue!8}
\textbf{Method} & \textbf{BPW} & \textbf{MMLU-R}$^*$ &
\textbf{GSM8K} & \textbf{MATH-500} & \textbf{IFEval} &
\textbf{HumanE+}$^*$ & \textbf{BFCL-v3} & \textbf{Average} \\
\midrule
\multicolumn{9}{l}{\textit{Qwen3-4B-Instruct-2507}} \\
\cmidrule[0.3pt]{1-9}
BF16     & 16.00 & 81.28 & 94.01 & 93.80 & 84.66 & 82.32 & 59.94 & 82.67 \\
\hdashline
QuIP\#   & 2.00 & 64.82 & 70.96 & 34.40 & 61.00 & 32.32 & 39.74 & 50.54 \\
QTIP     & 2.00 & 72.11 & 81.73 & 61.00 & 74.49 & 55.49 & 51.33 & 66.03 \\
UniSVQ   & 2.00 & 42.89 & 15.62 & 2.40 & 24.21 & 4.27 & 29.73 & 19.85 \\
SPHQuant & 2.25 & 66.93 & 81.35 & 66.40 & 80.04 & 57.32 & 39.48 & 65.25 \\
SAGE     & 2.25 & \textbf{73.47} & \textbf{88.70} & \textbf{78.80} & \textbf{82.81} & \textbf{59.76} & \textbf{53.33} & \textbf{72.81} \\
\midrule
\multicolumn{9}{l}{\textit{Qwen3-30B-A3B-Instruct-2507}} \\
\cmidrule[0.3pt]{1-9}
BF16     & 16.00 & 88.81 & 96.59 & 96.80 & 83.92 & 89.63 & 63.31 & 86.51 \\
\hdashline
QuIP\#   & 2.00 & 85.35 & 93.18 & 90.80 & 77.63 & 81.10 & 51.99 & 80.01 \\
QTIP     & 2.00 & \textbf{86.65} & 94.16 & 91.40 & 80.22 & 84.76 & \textbf{56.65} & 82.31 \\
UniSVQ   & 2.00 & 77.42 & 83.85 & 66.00 & 60.63 & 62.80 & 36.18 & 64.48 \\
SPHQuant & 2.25 & 83.18 & 93.18 & 88.40 & 80.41 & 75.61 & 49.79 & 78.43 \\
SAGE     & 2.25 & 85.86 & \textbf{94.31} & \textbf{93.60} & \textbf{80.96} & \textbf{85.37} & 54.77 & \textbf{82.48} \\
\midrule
\multicolumn{9}{l}{\textit{Qwen3.5-4B-Thinking}} \\
\cmidrule[0.3pt]{1-9}
BF16     & 16.00 & 89.19 & 96.89 & 93.20 & 87.62 & 83.54 & 64.19 & 85.77 \\
\hdashline
QuIP\#   & 2.00 & 36.98 & 40.18 & 15.80 & 27.17 & 13.41 & 24.13 & 26.28 \\
QTIP     & 2.00 & 68.60 & 76.80 & 25.80 & 49.17 & 28.66 & 41.93 & 48.49 \\
UniSVQ   & 2.00 & 7.53 & 0.68 & 0.00 & 1.48 & 0.00 & 28.85 & 6.42 \\
SPHQuant & 2.25 & 68.04 & 70.66 & 32.40 & 59.89 & 39.02 & 38.60 & 51.44 \\
SAGE     & 2.25 & \textbf{78.04} & \textbf{88.93} & \textbf{53.60} & \textbf{71.53} & \textbf{65.85} & \textbf{48.12} & \textbf{67.68} \\
\midrule
\multicolumn{9}{l}{\textit{Ministral-3-8B-Instruct-2512}} \\
\cmidrule[0.3pt]{1-9}
BF16     & 16.00 & 82.14 & 94.47 & 87.40 & 62.29 & 76.22 & 51.97 & 75.75 \\
\hdashline
QuIP\#   & 2.00 & 67.88 & 73.16 & 40.40 & 42.51 & 43.29 & 30.19 & 49.57 \\
QTIP     & 2.00 & \textbf{72.05} & 78.32 & 50.00 & 47.50 & 56.71 & 29.81 & 55.73 \\
UniSVQ   & 2.00 & 50.68 & 28.05 & 6.00 & 29.39 & 10.37 & 29.73 & 25.70 \\
SPHQuant & 2.25 & 63.65 & 67.55 & 44.80 & 48.43 & 50.00 & 35.01 & 51.57 \\
SAGE     & 2.25 & \textbf{72.05} & \textbf{84.23} & \textbf{63.40} & \textbf{53.42} & \textbf{57.32} & \textbf{48.85} & \textbf{63.21} \\
\midrule
\multicolumn{9}{l}{\textit{Ministral-3-14B-Instruct-2512}} \\
\cmidrule[0.3pt]{1-9}
BF16     & 16.00 & 84.82 & 95.75 & 90.20 & 64.88 & 81.10 & 56.50 & 78.88 \\
\hdashline
QuIP\#   & 2.00 & 73.65 & 85.97 & 55.60 & 52.68 & 57.32 & 39.00 & 60.70 \\
QTIP     & 2.00 & 70.74 & 83.09 & 46.80 & 50.28 & 30.49 & 29.71 & 51.85 \\
UniSVQ   & 2.00 & 63.30 & 64.37 & 27.20 & 46.40 & 26.83 & 28.91 & 42.84 \\
SPHQuant & 2.25 & 72.33 & 82.41 & 62.00 & \textbf{63.03} & 65.85 & 47.81 & 65.57 \\
SAGE     & 2.25 & \textbf{78.81} & \textbf{90.14} & \textbf{77.40} & 62.48 & \textbf{68.90} & \textbf{54.03} & \textbf{71.96} \\
\bottomrule
\end{tabular}%
}
\par\vspace{2pt}
{\footnotesize\raggedright
$^*$MMLU-R denotes MMLU-Redux, and HumanE+ denotes HumanEval+.\par}
\vspace{-5mm}
\end{table}

\vspace{-2mm}
\subsection{Experimental Settings}
\vspace{-2mm}
\paragraph{Implementation Details.}
Accuracy experiments use PyTorch~\citep{paszke2019pytorch} and HuggingFace Transformers~\citep{wolf2020transformers} on NVIDIA RTX PRO 6000 Blackwell GPUs; the inference-efficiency experiments in Sec.~\ref{sec:inference-efficiency} use an RTX 4090.
We perform blockwise optimization by minimizing the mean squared error between quantized and full-precision block outputs.
The radius and codebook parameters are jointly optimized using Adam for 20 epochs, with a batch size of 16 and a constant learning rate of $10^{-3}$.
The calibration corpus contains 1920 training sequences and 128 validation sequences, each with 2048 tokens.
We use a subvector dimension of $v=8$ and a radius group size of 128 weights.
We initialize codebooks with 20 iterations of activation-weighted spherical $k$-means on 65{,}536 sampled subvectors.
By default, ESG uses $\gamma=0.75$, and ABS selects $(K,V)\in\{(64,4),(128,3),(256,2)\}$ per matrix, unless otherwise stated.
\paragraph{Models and Evaluation.}
We evaluate five models covering dense, mixture-of-experts (MoE), and hybrid attention architectures, as well as instruct and thinking modes.
These include Qwen3-4B-Instruct-2507~\citep{qwen3}, Qwen3-30B-A3B-Instruct-2507 (MoE)~\citep{qwen3}, Ministral-3-8B-Instruct-2512~\citep{ministral3}, and Ministral-3-14B-Instruct-2512~\citep{ministral3} in instruct mode, and Qwen3.5-4B~\citep{qwen3.5}, which combines linear and full attention, in thinking mode.
For multimodal checkpoints, we quantize the language backbone.
We assess practical capabilities on six benchmarks: MMLU-Redux~\citep{mmlu-redux} for general knowledge; GSM8K~\citep{gsm8k} and MATH-500~\citep{math500} for mathematical reasoning; IFEval~\citep{ifeval} for instruction following; HumanEval+~\citep{humanevalplus} for code generation; and BFCL-v3~\citep{bfcl} for tool use.
We report accuracy for MMLU-Redux, GSM8K, and MATH-500, prompt-level strict accuracy for IFEval, pass@1 for HumanEval+, and the overall score for BFCL-v3. See Appendix for quantization and evaluation details.

\vspace{-3mm}
\paragraph{Baselines.}
We compare SAGE with four 2-bit weight-only VQ baselines: SPHQuant~\citep{sphquant}, using spherical quantization; QTIP~\citep{qtip}, using trellis coding with incoherence processing; QuIP\#~\citep{quip-sharp}, using lattice codebooks with Hadamard incoherence processing; and UniSVQ~\citep{unisvq}, unifying scalar and vector quantization through affine lattice codebooks. Speed benchmarks additionally include AWQ~\citep{awq} at W4G128.

\vspace{-2mm}
\subsection{Main Results}
\vspace{-2mm}

\paragraph{Comparisons with State-of-the-Art Methods.}
As shown in Table~\ref{tab:main-results}, SAGE achieves the highest average score on all five models, retaining 78.9--95.3\% of their BF16 average scores.
At the same 2.25-bit encoding budget, it improves over SPHQuant by 4.05--16.24 percentage points.
The largest gain occurs on Qwen3.5-4B-Thinking, where SAGE raises the average score from 51.44 to 67.68 and leads all quantized baselines on every benchmark.
On the three dense instruct models, SAGE surpasses the strongest baseline for each model by 6.39--7.48 points.
On Qwen3-30B-A3B, it achieves an average score of 82.48, retaining 95.3\% of BF16 performance. These results demonstrate gains across dense, MoE, and hybrid attention models in instruct and thinking modes.

\vspace{-2mm}
\paragraph{Task Sensitivity to Quantization.}
Quantization affects tasks unevenly, with code generation and challenging mathematical reasoning showing substantial losses.
For example, SAGE trails BF16 by 22.56 points on HumanEval+ but only 1.85 points on IFEval for Qwen3-4B, indicating that strong instruction following can coexist with substantial degradation in code generation.
Sensitivity also varies within mathematical reasoning.
On Qwen3.5-4B-Thinking, the gap to BF16 reaches 39.60 points on MATH-500, compared with 7.96 points on GSM8K.
This contrast highlights the difficulty of preserving demanding reasoning in a thinking model, even when performance on simpler mathematical tasks remains relatively strong.
Across five models, SAGE gains 2.44--26.83 points on HumanEval+ and 5.20--21.20 on MATH-500 over SPHQuant.
Remaining gaps to BF16 motivate task-level evaluation beyond average scores.

\begin{table}[t]
\centering
\caption{ESG and ABS ablation on Qwen3-4B-Instruct-2507. Best results are in \textbf{bold}.}
\label{tab:component-ablation}
\small
\setlength{\tabcolsep}{4pt}
\renewcommand{\arraystretch}{1.08}
\resizebox{\textwidth}{!}{%
\begin{tabular}{cc*{7}{c}}
\toprule
\rowcolor{blue!8}
\textbf{ESG} & \textbf{ABS} &
\textbf{MMLU-R}$^*$ & \textbf{GSM8K} & \textbf{MATH-500} &
\textbf{IFEval} & \textbf{HumanE+}$^*$ & \textbf{BFCL-v3} &
\textbf{Average} \\
\midrule
\ding{55} & \ding{55} &
66.93 & 81.35 & 66.40 & 80.04 & 57.32 & 39.48 & 65.25 \\
\ding{51} & \ding{55} &
69.05 & 82.18 & 68.40 & 80.04 & 54.88 & 45.17 & 66.62 \\
\ding{55} & \ding{51} &
71.33 & 84.84 & 74.40 & 79.85 & 58.54 & 44.02 & 68.83 \\
\ding{51} & \ding{51} &
\textbf{73.47} & \textbf{88.70} & \textbf{78.80} &
\textbf{82.81} & \textbf{59.76} & \textbf{53.33} & \textbf{72.81} \\
\bottomrule
\end{tabular}%
}
\par\vspace{2pt}
{\footnotesize\raggedright
$^*$MMLU-R denotes MMLU-Redux, and HumanE+ denotes HumanEval+.\par}
\vspace{-5mm}
\end{table}

\vspace{-2mm}
\subsection{Ablation Studies}
\vspace{-2mm}

\paragraph{Contributions of ESG and ABS.}
Table~\ref{tab:component-ablation} evaluates ESG and ABS on Qwen3-4B-Instruct-2507.
ESG alone improves the average score by 1.37 points, from 65.25 to 66.62, while ABS alone yields a 3.58-point gain, reaching 68.83.
Full SAGE achieves 72.81, improving over the baseline by 7.56 points and outperforming the individual-component configurations by 6.19 and 3.98 points, respectively.
It also achieves the best score across all six benchmarks, indicating broad improvements in capability preservation.
The larger gain of the full configuration suggests that grouping and allocation work complementarily.
ESG changes the structure and reconstruction priorities of subvectors, while ABS adapts each matrix's direction--radius split to this structure.
By providing a more suitable allocation of representational capacity at initialization, ABS can help realize the benefits of ESG.

\begin{figure}[t]
    \centering
    \includegraphics[width=\textwidth]{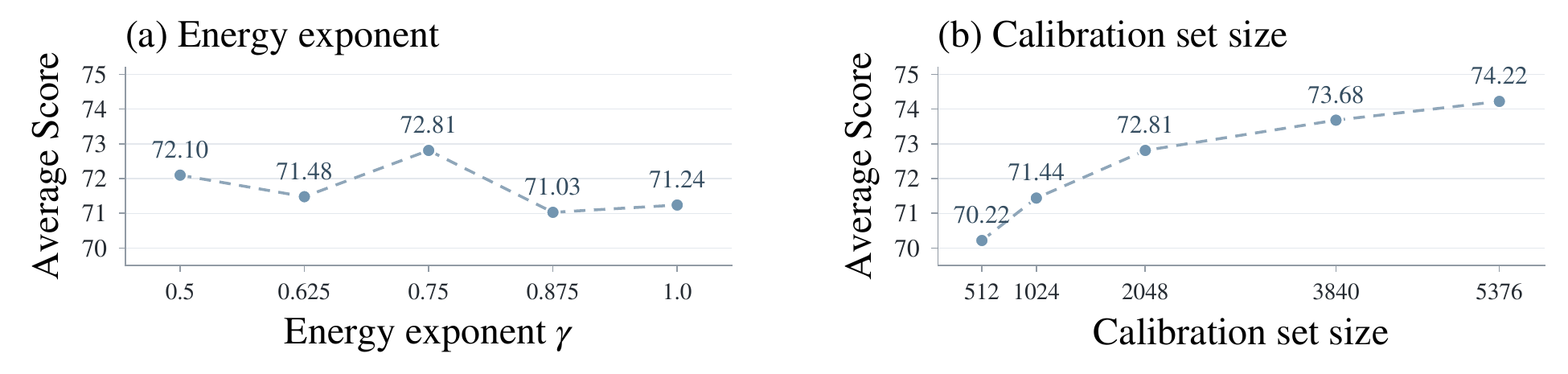}
    \vspace{-9mm}
    \caption{Sensitivity to (a) the energy exponent $\gamma$ and (b) calibration set size on Qwen3-4B-Instruct-2507. Scores are averaged over six benchmarks covering diverse generative capabilities.}
    \label{fig:sensitivity}
    \vspace{-6mm}
\end{figure}

\paragraph{Sensitivity to the Energy Exponent.}
Fig.~\ref{fig:sensitivity}(a) evaluates $\gamma$ from 0.5 to 1.0 on Qwen3-4B-Instruct-2507.
Average scores range from 71.03 to 72.81, with $\gamma=0.75$ achieving the highest score.
All tested settings outperform the baseline score of 65.25, suggesting that the benefits are maintained across a range of weighting strengths.
The exponent controls how strongly activation-energy differences influence reconstruction: smaller values produce more uniform weights, while larger values emphasize high-energy channels.
We default to $\gamma=0.75$, which achieves the highest average score.

\vspace{-2mm}
\paragraph{Effect of Calibration Set Size.}
As shown in Fig.~\ref{fig:sensitivity}(b), increasing the calibration set size from 512 to 5376 sequences improves SAGE's average score from 70.22 to 74.22.
The largest setting outperforms our default by 1.41 percentage points, suggesting that additional calibration data can further improve capability preservation.
Given the computational cost of larger calibration sets, we use the same calibration set size as SPHQuant in main experiments to match the data budget.

\begin{figure}[h]
\vspace{-2mm}
\centering
\begingroup
\fontsize{14bp}{16bp}\selectfont
\setlength{\tabcolsep}{5bp}
\renewcommand{\arraystretch}{0.98}
\resizebox{\textwidth}{!}{%
\includegraphics{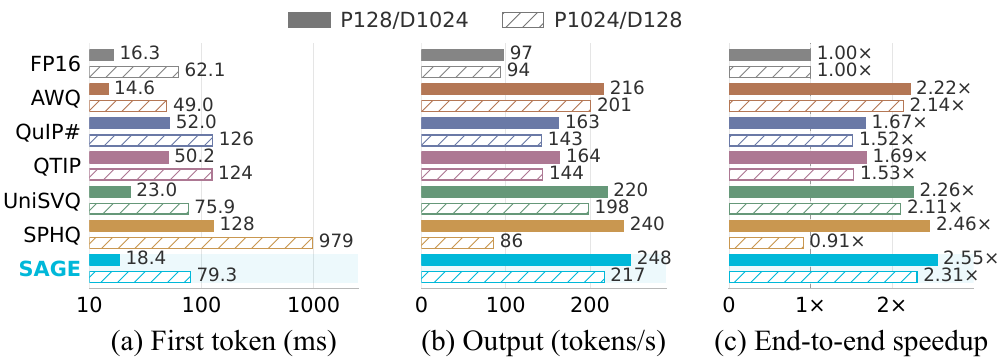}%
\hspace{14bp}%
\raisebox{4.46149bp}{%
\begin{tabular}[b]{@{}c@{}}
\begin{tabular}[b]{@{}ccc@{}}
\toprule
\textbf{Method} & \textbf{4B} & \textbf{30B-A3B} \\
\midrule
FP16 & 7.49 & 56.87 \\
QuIP\# & 1.58\textsubscript{4.76×↓} & 8.24\textsubscript{6.90×↓} \\
QTIP & 1.58\textsubscript{4.75×↓} & 8.32\textsubscript{6.84×↓} \\
UniSVQ & 1.57\textsubscript{4.76×↓} & 8.24\textsubscript{6.90×↓} \\
SPHQ & 1.78\textsubscript{4.20×↓} & 9.90\textsubscript{5.74×↓} \\
SAGE & 1.79\textsubscript{4.19×↓} & 10.02\textsubscript{5.68×↓} \\
\bottomrule
\end{tabular} \\[14bp]
(d) Packed model size (GiB)
\end{tabular}}%
}
\endgroup
\vspace{-5mm}
\caption{\textbf{Inference efficiency.}
Qwen3-4B-Instruct-2507, RTX 4090 (non-streaming, batch 1).
P/D: input/output tokens. Output throughput includes prefill; weights are reconstructed on demand. Packed model sizes include unquantized weights; AWQ uses W4G128 for speed benchmarks.}
\label{fig:inference-efficiency}
\vspace{-4mm}
\end{figure}

\subsection{Inference Efficiency}
\label{sec:inference-efficiency}
\vspace{-2mm}

\paragraph{End-to-End Inference Speed.}
We benchmark non-streaming, batch-one greedy generation with packed weights and FP16 activations and KV cache.
As shown in Fig.~\ref{fig:inference-efficiency}, SAGE achieves a \SAGEMainSpeedupRange$\times$ end-to-end speedup over dense FP16, the highest throughput in both workloads, and \SAGEFirstTokenReductionRange\% lower first-token latency than SPHQuant.
Detailed protocols and the tradeoff between latency and memory usage when retaining reconstructed weights are provided in Appendix~\ref{sec:inference-protocol}.
\vspace{-2mm}
\paragraph{Memory and Storage Costs.}
Following prior work~\citep{sphquant,qtip,unisvq}, we quantize all linear layers except the LM head, leaving embeddings and LayerNorm parameters unquantized.
SAGE peaks at 2.19\,GiB versus 7.94\,GiB for FP16 (a $3.63\times$ reduction).
The encoding payload uses 2.25 bits per weight, with codebook and permutation metadata stored separately.
Fig.~\ref{fig:inference-efficiency}(d) reports packed model sizes, including unquantized parameters.

\vspace{-2mm}
\section{Conclusion}
\vspace{-2mm}
In this work, we presented SAGE, a weight-only vector quantization method that uses activation information to guide channel grouping and bit allocation.
These two decisions are closely connected: grouping determines which weights share a representation, while bit allocation determines its capacity to encode direction and magnitude.
By addressing them together, SAGE makes more effective use of a limited encoding budget.
Experiments across diverse generative benchmarks demonstrate improved capability preservation, while our custom CUDA kernel translates the compact representation into practical inference acceleration.
We hope this work helps move extreme low-bit quantization beyond evaluations centered on perplexity toward models that reliably solve real-world tasks.

\bibliography{iclr2027_conference}

@book{gersho,
    title={Vector quantization and signal compression},
    author={Gersho, Allen and Gray, Robert M},
    year={1991},
    publisher={Springer New York, NY}
}

@article{ayot,
    title={Attend to Your Own Thoughts: Breaking the Barrier for Post-Training Quantization of Reasoning LLMs through the Lens of 1.58-Bit Quantization},
    author={Wang, Shigeng and Li, Chao and Kang, Yangyuxuan and Fan, Jiawei and Yao, Anbang},
    journal={arXiv preprint arXiv:2608.01078},
    year={2026}
}

@article{sphquant,
    title={{SPHQuant}: Efficient extreme low bit weight quantization for Vision-Language Models}, 
    author={Kewei Zhang and Zheng Chen and Haotong Qin and Yulun Zhang},
    journal={NeurIPS},
    year={2026}
}

@inproceedings{paszke2019pytorch,
    title={PyTorch: An Imperative Style, High-Performance Deep Learning Library.},
    author={Adam Paszke and Sam Gross and Francisco Massa and Adam Lerer and James Bradbury and Gregory Chanan and Trevor Killeen and Zeming Lin and Natalia Gimelshein and Luca Antiga and Alban Desmaison and Andreas Köpf and Edward Yang and Zachary DeVito and Martin Raison and Alykhan Tejani and Sasank Chilamkurthy and Benoit Steiner and Lu Fang and Junjie Bai and Soumith Chintala},
    booktitle={NeurIPS},
    year={2019}
}

@inproceedings{wolf2020transformers,
    title={Transformers: State-of-the-Art Natural Language Processing},
    author={Thomas Wolf and Lysandre Debut and Victor Sanh and Julien Chaumond and Clement Delangue and Anthony Moi and Pierric Cistac and Tim Rault and Rémi Louf and Morgan Funtowicz and Joe Davison and Sam Shleifer and Patrick von Platen and Clara Ma and Yacine Jernite and Julien Plu and Canwen Xu and Teven Le Scao and Sylvain Gugger and Mariama Drame and Quentin Lhoest and Alexander M. Rush},
    booktitle={EMNLP (Demos)},
    year={2020}
}

@article{gsm8k,
    title={Training Verifiers to Solve Math Word Problems},
    author={Cobbe, Karl and Kosaraju, Vineet and Bavarian, Mohammad and Chen, Mark and Jun, Heewoo and Kaiser, Lukasz and Plappert, Matthias and Tworek, Jerry and Hilton, Jacob and Nakano, Reiichiro and Hesse, Christopher and Schulman, John},
    journal={arXiv preprint arXiv:2110.14168},
    year={2021}
}

@inproceedings{bfcl,
    title={The Berkeley Function Calling Leaderboard ({BFCL}): From Tool Use to Agentic Evaluation of Large Language Models},
    author={Shishir G Patil and Huanzhi Mao and Fanjia Yan and Charlie Cheng-Jie Ji and Vishnu Suresh and Ion Stoica and Joseph E. Gonzalez},
    booktitle={ICML},
    year={2025},
}

@inproceedings{math500,
    title={Let's Verify Step by Step},
    author={Hunter Lightman and Vineet Kosaraju and Yuri Burda and Harrison Edwards and Bowen Baker and Teddy Lee and Jan Leike and John Schulman and Ilya Sutskever and Karl Cobbe},
    booktitle={ICLR},
    year={2024}
}

@inproceedings{humanevalplus,
    title={Is Your Code Generated by Chat{GPT} Really Correct? Rigorous Evaluation of Large Language Models for Code Generation},
    author={Liu, Jiawei and Xia, Chunqiu Steven and Wang, Yuyao and ZHANG, LINGMING},
    booktitle={NeurIPS},
    year={2023}
}

@inproceedings{mmlu-redux,
    title={Are We Done with MMLU?},
    author={Aryo Pradipta Gema and Joshua Ong Jun Leang and Giwon Hong and Alessio Devoto and Alberto Carlo Maria Mancino and Rohit Saxena and Xuanli He and Yu Zhao and Xiaotang Du and Mohammad Reza Ghasemi Madani and Claire Barale and Robert McHardy and Joshua Harris and Jean Kaddour and Emile van Krieken and Pasquale Minervini},
    booktitle={NAACL (Long Papers)},
    year={2025}
}

@article{ifeval,
  title={Instruction-Following Evaluation for Large Language Models},
  author={Zhou, Jeffrey and Lu, Tianjian and Mishra, Swaroop and Brahma, Siddhartha and Basu, Sujoy and Luan, Yi and Zhou, Denny and Hou, Le},
  journal={arXiv preprint arXiv:2311.07911},
  year={2023}
}

@misc{qwen3.5,
    title  = {{Qwen3.5}: Towards Native Multimodal Agents},
    author = {{Qwen Team}},
    month  = {February},
    year   = {2026},
    url    = {https://qwen.ai/blog?id=qwen3.5}
}

@article{qwen3,
    title={Qwen3 technical report},
    author={Yang, An and Li, Anfeng and Yang, Baosong and Zhang, Beichen and Hui, Binyuan and Zheng, Bo and Yu, Bowen and Gao, Chang and Huang, Chengen and Lv, Chenxu and others},
    journal={arXiv preprint arXiv:2505.09388},
    year={2025}
}

@article{dsv4,
    title={Deepseek-v4: Towards highly efficient million-token context intelligence},
    author={{DeepSeek Team}},
    journal={arXiv preprint arXiv:2606.19348},
    year={2026}
}

@article{ministral3,
    title={Ministral 3},
    author={Liu, Alexander H and Khandelwal, Kartik and Subramanian, Sandeep and Jouault, Victor and Rastogi, Abhinav and Sad{\'e}, Adrien and Jeffares, Alan and Jiang, Albert and Cahill, Alexandre and Gavaudan, Alexandre and others},
    journal={arXiv preprint arXiv:2601.08584},
    year={2026}
}

@article{kimi-k3,
    title={{Kimi K3}: Open Frontier Intelligence},
    author={{Kimi Team}},
    journal={arXiv preprint arXiv:2607.24653},
    year={2026}
}

@article{glm5,
    title={{GLM-5}: from Vibe Coding to Agentic Engineering},
    author={{GLM-5 Team}},
    journal={arXiv preprint arXiv:2602.15763},
    year={2026}
}

@inproceedings{vllm,
    title={Efficient Memory Management for Large Language Model Serving with PagedAttention},
    author={Woosuk Kwon and Zhuohan Li and Siyuan Zhuang and Ying Sheng and Lianmin Zheng and Cody Hao Yu and Joseph Gonzalez and Hao Zhang and Ion Stoica},
    booktitle={SOSP},
    year={2023}
}

@inproceedings{powerinfer,
    title={Powerinfer: Fast large language model serving with a consumer-grade gpu},
    author={Song, Yixin and Mi, Zeyu and Xie, Haotong and Chen, Haibo},
    booktitle={SOSP},
    year={2024}
}

@inproceedings{paroquant,
    title={ParoQuant: Pairwise Rotation Quantization for Efficient Reasoning {LLM} Inference},
    author={Yesheng Liang and Haisheng Chen and Song Han and Zhijian Liu},
    booktitle={ICLR},
    year={2026}
}

@article{pcdvq,
    title={Pcdvq: Enhancing vector quantization for large language models via polar coordinate decoupling},
    author={Yue, Yuxuan and Xu, Zukang and Yuan, Zhihang and Yang, Dawei and Wu, Jianlong and Nie, Liqiang},
    journal={arXiv preprint arXiv:2506.05432},
    year={2025}
}

@inproceedings{llmint8,
    title={Gpt3. int8 (): 8-bit matrix multiplication for transformers at scale},
    author={Dettmers, Tim and Lewis, Mike and Belkada, Younes and Zettlemoyer, Luke},
    booktitle={NeurIPS},
    year={2022}
}

@inproceedings{smoothquant,
    title={SmoothQuant: Accurate and Efficient Post-Training Quantization for Large Language Models},
    author={Guangxuan Xiao and Ji Lin and Mickaël Seznec and Hao Wu and Julien Demouth and Song Han},
    booktitle={ICML},
    year={2023}
}

@inproceedings{gptq,
    title={{OPTQ}: Accurate Quantization for Generative Pre-trained Transformers},
    author={Elias Frantar and Saleh Ashkboos and Torsten Hoefler and Dan Alistarh},
    booktitle={ICLR},
    year={2023}
}

@inproceedings{awq,
    title={Awq: Activation-aware weight quantization for on-device llm compression and acceleration},
    author={Lin, Ji and Tang, Jiaming and Tang, Haotian and Yang, Shang and Chen, Wei-Ming and Wang, Wei-Chen and Xiao, Guangxuan and Dang, Xingyu and Gan, Chuang and Han, Song},
    booktitle={MLSys},
    year={2024}
}

@inproceedings{squeezellm,
    title={Squeeze{LLM}: Dense-and-Sparse Quantization},
    author={Sehoon Kim and Coleman Richard Charles Hooper and Amir Gholami and Zhen Dong and Xiuyu Li and Sheng Shen and Michael W. Mahoney and Kurt Keutzer},
    booktitle={ICML},
    year={2024}
}

@inproceedings{omniquant,
    title={OmniQuant: Omnidirectionally Calibrated Quantization for Large Language Models},
    author={Wenqi Shao and Mengzhao Chen and Zhaoyang Zhang and Peng Xu and Lirui Zhao and Zhiqian Li and Kaipeng Zhang and Peng Gao and Yu Qiao and Ping Luo},
    booktitle={ICLR},
    year={2024}
}

@inproceedings{quip,
  title={Quip: 2-bit quantization of large language models with guarantees},
  author={Jerry Chee and Yaohui Cai and Volodymyr Kuleshov and Christopher De Sa},
  booktitle={NeurIPS},
  year={2023}
}

@inproceedings{quarot,
    title={QuaRot: Outlier-Free 4-Bit Inference in Rotated LLMs},
    author={Saleh Ashkboos and Amirkeivan Mohtashami and Maximilian L. Croci and Bo Li and Pashmina Cameron and Martin Jaggi and Dan Alistarh and Torsten Hoefler and James Hensman},
    booktitle={NeurIPS},
    year={2024}
}

@inproceedings{spinquant,
    title={SpinQuant: {LLM} Quantization with Learned Rotations},
    author={Zechun Liu and Changsheng Zhao and Igor Fedorov and Bilge Soran and Dhruv Choudhary and Raghuraman Krishnamoorthi and Vikas Chandra and Yuandong Tian and Tijmen Blankevoort},
    booktitle={ICLR},
    year={2025}
}

@inproceedings{leanquant,
    title={LeanQuant: Accurate and Scalable Large Language Model Quantization with Loss-error-aware Grid},
    author={Tianyi Zhang and Anshumali Shrivastava},
    booktitle={ICLR},
    year={2025}
}

@inproceedings{llmqat,
    title={{LLM}-{QAT}: Data-Free Quantization Aware Training for Large Language Models},
    author={Zechun Liu and Barlas Oguz and Changsheng Zhao and Ernie Chang and Pierre Stock and Yashar Mehdad and Yangyang Shi and Raghuraman Krishnamoorthi and Vikas Chandra},
    booktitle={ACL (Findings)},
    year={2024}
}

@inproceedings{efficientqat,
    title={EfficientQAT: Efficient Quantization-Aware Training for Large Language Models},
    author={Mengzhao Chen and Wenqi Shao and Peng Xu and Jiahao Wang and Peng Gao and Kaipeng Zhang and Ping Luo},
    booktitle={ACL (1)},
    year={2025}
}

@inproceedings{paretoq,
    title={ParetoQ: Improving Scaling Laws in Extremely Low-bit {LLM} Quantization},
    author={Zechun Liu and Changsheng Zhao and Hanxian Huang and Sijia Chen and Jing Zhang and Jiawei Zhao and Scott Roy and Lisa Jin and Yunyang Xiong and Yangyang Shi and Lin Xiao and Yuandong Tian and Bilge Soran and Raghuraman Krishnamoorthi and Tijmen Blankevoort and Vikas Chandra},
    booktitle={NeurIPS},
    year={2025},
}

@inproceedings{qep,
    title={Quantization Error Propagation: Revisiting Layer-Wise Post-Training Quantization},
    author={Yamato Arai and Yuma Ichikawa},
    booktitle={NeurIPS},
    year={2025}
}

@inproceedings{bitdistiller,
    title={Bitdistiller: Unleashing the potential of sub-4-bit llms via self-distillation},
    author={Dayou Du and Yijia Zhang and Shijie Cao and Jiaqi Guo and Ting Cao and Xiaowen Chu and Ningyi Xu},
    booktitle={ACL (1)},
    year={2024}
}

@inproceedings{flatquant,
    title={FlatQuant: Flatness Matters for {LLM} Quantization},
    author={Yuxuan Sun and Ruikang Liu and Haoli Bai and Han Bao and Kang Zhao and Yuening Li and JiaxinHu and Xianzhi Yu and Lu Hou and Chun Yuan and Xin Jiang and Wulong Liu and Jun Yao},
    booktitle={ICML},
    year={2025}
}

@inproceedings{ostquant,
    title={OSTQuant: Refining Large Language Model Quantization with Orthogonal and Scaling Transformations for Better Distribution Fitting},
    author={Xing Hu and Yuan Cheng and Dawei Yang and Zhixuan Chen and Zukang Xu and Jiangyong Yu and Chen Xu and Zhihang Yuan and Zhe Jiang and Sifan Zhou},
    booktitle={ICLR},
    year={2025}
}

@inproceedings{gptaq,
    title={{GPTAQ}: Efficient Finetuning-Free Quantization for Asymmetric Calibration},
    author={Yuhang Li and Ruokai Yin and Donghyun Lee and Shiting Xiao and Priyadarshini Panda},
    booktitle={ICML},
    year={2025}
}

@inproceedings{gptvq,
    title={{GPTVQ}: The Blessing of Dimensionality for {LLM} Quantization},
    author={Mart Van Baalen and Andrey Kuzmin and Markus Nagel and Peter Couperus and Artem Bolshakov and Cedric Bastoul and Eric Mahurin and Tijmen Blankevoort and Paul Whatmough},
    booktitle={Workshop on Efficient Systems for Foundation Models II @ ICML2024},
    year={2024}
}

@inproceedings{aqlm,
    title={Extreme compression of large language models via additive quantization},
    author={Vage Egiazarian and Andrei Panferov and Denis Kuznedelev and Elias Frantar and Artem Babenko and Dan Alistarh},
    booktitle={ICML},
    year={2024}
}

@article{quip-sharp,
    title={Qu{IP}\#: Even Better {LLM} Quantization with Hadamard Incoherence and Lattice Codebooks},
    author={Albert Tseng and Jerry Chee and Qingyao Sun and Volodymyr Kuleshov and Christopher De Sa},
    journal={ICML},
    year={2024}
}

@inproceedings{pvtuning,
    title={{PV}-Tuning: Beyond Straight-Through Estimation for Extreme {LLM} Compression},
    author={Vladimir Malinovskii and Denis Mazur and Ivan Ilin and Denis Kuznedelev and Konstantin Pavlovich Burlachenko and Kai Yi and Dan Alistarh and Peter Richt{\'a}rik},
    booktitle={NeurIPS},
    year={2024}
}

@inproceedings{qtip,
    title={{QTIP}: Quantization with Trellises and Incoherence Processing},
    author={Albert Tseng and Qingyao Sun and David Hou and Christopher De Sa},
    booktitle={NeurIPS},
    year={2024}
}

@inproceedings{vptq,
    title={VPTQ: Extreme Low-bit Vector Post-Training Quantization for Large Language Models},
    author={Yifei Liu and Jicheng Wen and Yang Wang and Shengyu Ye and Li Lyna Zhang and Ting Cao and Cheng Li and Mao Yang},
    booktitle={EMNLP},
    year={2024}
}

@inproceedings{nestquant,
    title={NestQuant: nested lattice quantization for matrix products and {LLM}s},
    author={Semyon Savkin and Eitan Porat and Or Ordentlich and Yury Polyanskiy},
    booktitle={ICML},
    year={2025}
}

@article{mgvq,
    title={MGVQ: Synergizing Multi-dimensional Sensitivity-Aware and Gradient-Hessian Fusion for Vector Quantization},
    author={Wang, Zhong and Xu, Zukang and Hu, Xing and Yang, Dawei},
    journal={arXiv preprint arXiv:2605.24019},
    year={2026}
}

@inproceedings{unisvq,
    title={Uni{SVQ}: 2-bit Unified Scalar-Vector Quantization},
    author={Haoyu Wang and Haiyan Zhao and Xingyu Yu and Zhangyang Yao and Xu Han and Zhiyuan Liu and Maosong Sun},
    booktitle={ICML},
    year={2026}
}

@inproceedings{lcqat,
    title={{LC}-{QAT}: Data-Efficient 2-Bit {QAT} for {LLM}s via Linear-Constrained Vector Quantization},
    author={Haoyu Wang and Xingyu Yu and Haiyan Zhao and Fengxiang Wang and Xu Han},
    booktitle={ICML},
    year={2026}
}

@inproceedings{tileq,
    title={TileQ: Efficient Low-Rank Quantization of Mixture-of-Experts with 2D Tiling},
    author={Hongyaoxing Gu and Xinzhe Chen and Lijuan Hu and Liu fangfang},
    booktitle={ICML},
    year={2026}
}

@inproceedings{arbllm,
    title={{ARB}-{LLM}: Alternating Refined Binarizations for Large Language Models},
    author={Zhiteng Li and Xianglong Yan and Tianao Zhang and Haotong Qin and Dong Xie and Jiang Tian and zhongchao shi and Linghe Kong and Yulun Zhang and Xiaokang Yang},
    booktitle={ICLR},
    year={2025}
}

@inproceedings{pt2llm,
    title={{PT}$^2$-{LLM}: Post-Training Ternarization for Large Language Models},
    author={Xianglong Yan and ChengZhu Bao and Zhiteng Li and Tianao Zhang and Kaicheng Yang and Haotong Qin and Ruobing Xie and Xingwu Sun and Yulun Zhang},
    booktitle={ICLR},
    year={2026}
}

@inproceedings{llvq,
    title={Leech Lattice Vector Quantization for Efficient {LLM} Compression},
    author={Tycho F. A. van der Ouderaa and Mart Van Baalen and Paul N. Whatmough and Markus Nagel},
    booktitle={AdaptFM: Resource-Adaptive Foundation Model Inference},
    year={2026}
}

@inproceedings{hellaswag,
    title={Hellaswag: Can a machine really finish your sentence?},
    author={Rowan Zellers and Ari Holtzman and Yonatan Bisk and Ali Farhadi and Yejin Choi},
    booktitle={ACL (1)},
    year={2019}
}

@article{opencodeinstruct,
  title={Opencodeinstruct: A large-scale instruction tuning dataset for code llms},
  author={Ahmad, Wasi Uddin and Ficek, Aleksander and Samadi, Mehrzad and Huang, Jocelyn and Noroozi, Vahid and Majumdar, Somshubra and Ginsburg, Boris},
  journal={arXiv preprint arXiv:2504.04030},
  year={2025}
}

@inproceedings{deepmath,
  title={Deepmath-103k: A large-scale, challenging, decontaminated, and verifiable mathematical dataset for advancing reasoning},
  author={Zhiwei He and Tian Liang and Jiahao Xu and Qiuzhi Liu and Xingyu Chen and Yue Wang and Linfeng Song and Dian Yu and Zhenwen Liang and Wenxuan Wang and Zhuosheng Zhang and Rui Wang and Zhaopeng Tu and Haitao Mi and Dong Yu},
  booktitle={ICLR},
  year={2026}
}

@inproceedings{ultrachat,
  title={Enhancing chat language models by scaling high-quality instructional conversations},
  author={Ning Ding and Yulin Chen and Bokai Xu and Yujia Qin and Shengding Hu and Zhiyuan Liu and Maosong Sun and Bowen Zhou},
  booktitle={EMNLP},
  year={2023}
}

@inproceedings{ross2025when2call,
  title={When2Call: When (not) to call tools},
  author={Ross, Hayley and Mahabaleshwarkar, Ameya Sunil and Suhara, Yoshi},
  booktitle={NAACL},
  year={2025}
}

@article{math,
  title={Measuring Mathematical Problem Solving With the {MATH} Dataset},
  author={Dan Hendrycks and Collin Burns and Saurav Kadavath and Akul Arora and Steven Basart and Eric Tang and Dawn Song and Jacob Steinhardt},
  journal={Thirty-fifth Conference on NeurIPS Datasets and Benchmarks Track (Round 2)},
  year={2021}
}
\bibliographystyle{iclr2027_conference}

\newpage
\appendix
\section{Appendix}

\subsection{CUDA Kernel Implementation Details}
\label{appendix_kernel_details}

\paragraph{GEMV.}
During decoding, each eight-weight subvector contributes a scalar to one output. Its radius is shared by all eight weights, so we apply that radius after taking the dot product with the signed direction (Fig.~\ref{fig:kernel-details}a). This replaces eight radius multiplications with one. In the shape-specialized kernels, we place each codeword's eight components consecutively in shared memory, allowing a thread to fetch the whole direction in one 128-bit load. We divide each 128-channel group among four neighboring threads. Each thread processes four subvectors, while other four-thread teams cover the remaining groups. Assigning multiple outputs to a thread lets it reuse the same activations; using fewer outputs per block instead increases the number of independent blocks.

\paragraph{GEMM.}
During prefill, many token rows use the same weights, so we reconstruct each weight once per output tile and reuse it across those rows. The schedule in Fig.~\ref{fig:kernel-details}b processes each 128-channel group in two halves. We decode both weight halves together from the same packed group and retain them until every warp finishes the group. Only the activation half currently being multiplied occupies shared memory, leaving room for other resident thread blocks. We fetch the next half's activations into registers before the current matrix multiply. A barrier prevents us from overwriting the shared activation buffer until every warp has finished reading it; another makes the replacement visible before the next multiply. Likewise, we prefetch the next group's packed weights while the current group is still in use, but wait to decode them into shared memory until both current weight halves have been consumed.

\begin{figure}[ht]
\centering
\begin{minipage}[t]{0.59\linewidth}
\vspace{0pt}
\raggedright
\fontsize{8}{10}\selectfont
\textbf{(a) Data paths within a thread block}\par
\vspace{4pt}
\centering
\begin{tikzpicture}[x=0.96cm,y=0.90cm,font=\fontsize{8}{9}\selectfont,
    box/.style={draw=black!65,rounded corners=1pt,line width=0.45pt,align=center,inner sep=3pt},
    shared/.style={box,fill=black!13},
    reg/.style={box,fill=white},
    flow/.style={-{Stealth[length=1.4mm]},line width=0.55pt}]
\def\kernelgap{0.083}
\path[use as bounding box] (0,0) rectangle (8.4,0);
\node[box,fill=black!5,minimum width=7.9cm,anchor=north] (global) at (4.2,0) {Global memory: packed codes, grid, codebook, input};
\node[anchor=west,font=\fontsize{8}{9}\selectfont\bfseries] at (0,{-0.87-\kernelgap}) {GEMV};
\node[shared,minimum width=3.05cm] (book) at (1.85,{-1.49-\kernelgap}) {Codebook in shared memory\\Once per block; contiguous $c_z$};
\node[reg,minimum width=3.8cm] (fields) at (6.2,{-1.49-\kernelgap}) {Registers: $s,z,q$\\Bit extraction from packed words};
\draw[flow] (global.south) -- ++(0,-0.2) -| (fields.north);
\draw[flow] (global.south) -- ++(0,-0.2) -| (book.north);
\node[reg,minimum width=7.9cm] (dot) at (4.2,{-2.68-2*\kernelgap}) {Registers: $(s\odot c_z)^\top x_{\mathrm{sub}}$\\Four \texttt{half2} FMAs; FP16 local accumulation};
\draw[flow] (book.south) -- (book.south |- dot.north);
\draw[flow] (fields.south) -- (fields.south |- dot.north);
\node[reg,minimum width=7.9cm] (radius) at (4.2,{-3.77-3*\kernelgap}) {FP32 radius weighting: $\hat r=\mu_g+q\delta_g$\\Four-lane shuffle sum};
\draw[flow] (dot) -- (radius);
\node[shared,minimum width=7.9cm] (reduce) at (4.2,{-4.66-4*\kernelgap}) {FP32 team partials $\longrightarrow$ sum $\longrightarrow$ FP16 output};
\draw[flow] (radius) -- (reduce);
\draw[black!30,line width=0.4pt] (0,{-5.08-4*\kernelgap}) -- (8.4,{-5.08-4*\kernelgap});
\node[anchor=west,font=\fontsize{8}{9}\selectfont\bfseries] at (0,{-5.41-4*\kernelgap}) {GEMM: $64\times64$ output tile};
\node[reg,minimum width=7.9cm] (reconstruct) at (4.2,{-6.01-4*\kernelgap}) {Registers: $\hat r(s\odot c_z)$ in FP32 $\longrightarrow$ FP16};
\node[shared,minimum width=3.05cm] (a) at (1.85,{-7.05-5*\kernelgap}) {Activation buffer $A$\\$64\times64$; reused per slab};
\node[shared,minimum width=3.8cm] (b) at (6.2,{-7.05-5*\kernelgap}) {Weight buffers $B_0,B_1$\\Two $64\times64$ slabs};
\draw[flow] (reconstruct.south -| b.north) -- (b.north);
\node[reg,minimum width=7.9cm] (mma) at (4.2,{-8.19-6*\kernelgap}) {Tensor Core: $A B_j^\top$\\FP16 operands; FP32 register accumulators};
\draw[flow] (a.south) -- (a.south |- mma.north);
\draw[flow] (b.south) -- (b.south |- mma.north);
\node[shared,minimum width=7.9cm] (epilogue) at (4.2,{-9.13-7*\kernelgap}) {FP32 epilogue $\longrightarrow$ FP16 output or FP32 split partials};
\draw[flow] (mma) -- (epilogue);
\end{tikzpicture}
\end{minipage}\hfill
\begin{minipage}[t]{0.39\linewidth}
\vspace{0pt}
\fontsize{8}{10}\selectfont
\textbf{(b) GEMM slab schedule}\par
\vspace{4pt}
$g$ starts at this block's first group. The FP32 accumulator \texttt{acc} starts at zero; the codebook is already in shared memory.\par
\vspace{6pt}
\begingroup
\fontsize{7.5}{11.5}\selectfont\ttfamily
\setlength{\tabcolsep}{0pt}
\begin{tabular}{@{}l@{}}
stage\_A(load\_A(g,0))\\
stage\_B(decode(load\_Q(g)))\\
sync()\\
for g in assigned\_groups:\\
\quad a1 = load\_A(g,1)\\
\quad mma(A,B[0],acc)\\
\quad sync(); stage\_A(a1)\\
\quad if has\_next(g):\\
\qquad a0 = load\_A(g+1,0)\\
\qquad q1 = load\_Q(g+1)\\
\quad sync(); mma(A,B[1],acc)\\
\quad sync()\\
\quad if has\_next(g):\\
\qquad stage\_A(a0)\\
\qquad stage\_B(decode(q1))\\
\quad sync()
\end{tabular}
\endgroup
\par\vspace{6pt}
\raggedright
\texttt{load\_A/Q} reads activations or codes and grid into registers. \texttt{stage\_A/B} writes shared memory. \texttt{decode} reconstructs both weight halves; \texttt{sync} waits for all threads.\par
\end{minipage}
\caption{Kernel data paths and the 64-row GEMM loop. $x_{\mathrm{sub}}$ contains eight reordered activations. In (b), $j\in\{0,1\}$ selects a half-group, and \texttt{mma} adds $A B_j^\top$ to \texttt{acc}.}
\label{fig:kernel-details}
\end{figure}

\subsection{Inference Measurement Details}
\label{sec:inference-protocol}
\paragraph{Hardware and preparation.}
We benchmark Qwen3-4B-Instruct-2507 on one RTX 4090 with 14 AMD EPYC 9354 CPU cores and 60.1\,GB RAM, using Python 3.11, PyTorch 2.10.0 (cu128), CUDA 12.8, cuDNN 9.10.2, and Transformers 5.15.0.
Unlike accuracy checkpoints, timing checkpoints use synthetic packed weights, SAGE permutations, and bit splits; only decoder linear layers are replaced.
Batch size is one; activations and the static KV cache are FP16, with cache capacity $P+D-1$ for $P$ input and $D$ output tokens.
UniSVQ uses BF16 internally, including boundary casts in timing.
SDPA priority is cuDNN, FlashAttention, memory-efficient attention, then math, during both compilation and execution; an untimed profiler checks actual kernel selection.
Prefill and decode use \texttt{torch.compile} with \texttt{fullgraph=True}, \texttt{dynamic=False}, and \texttt{reduce-overhead}: two compilation/capture passes, profiling, one recovery pass, five warmups, then ten measurements with recompilation forbidden.
Loading, input preparation, compilation, profiling, and per-window setup (cache reset, pinned buffers, events, copy stream, observer, and device synchronization) are untimed.

\begin{figure}[ht]
\centering
\resizebox{0.8\linewidth}{!}{
\begin{tikzpicture}[x=0.95cm,y=0.85cm,font=\small,
    axis/.style={-{Stealth[length=2mm]},line width=0.7pt},
    dependency/.style={-{Stealth[length=2mm]},line width=0.7pt},
    interval/.style={{Stealth[length=1.7mm]}-{Stealth[length=1.7mm]},line width=0.6pt}]
\node[anchor=east,font=\small\bfseries] at (0.85,1.35) {CPU};
\node[anchor=east,font=\small\bfseries] at (0.85,0) {GPU};
\draw[axis] (1.05,1.35) -- (13.1,1.35);
\draw[axis] (1.05,0) -- (13.1,0);
\draw[fill=black!8,line width=0.6pt] (1.85,-0.26) rectangle (4.05,0.26);
\draw[fill=black!8,line width=0.6pt] (4.05,-0.26) rectangle (11.95,0.26);
\node at (2.95,0) {Prefill};
\node at (8,0) {Decode: $D-1$ tokens};
\draw[dependency] (1.5,1.35) -- (1.85,0.26);
\draw[dependency] (4.05,0.26) -- (4.65,1.35);
\draw[dependency] (11.95,0.26) -- (12.3,1.35);
\draw[line width=0.5pt] (1.5,1.35) -- (1.5,2.5);
\draw[line width=0.5pt] (4.65,1.35) -- (4.65,1.93);
\draw[line width=0.5pt] (12.3,1.35) -- (12.3,2.5);
\draw[interval] (1.5,2.4) -- node[above,inner sep=2pt] {Total $=t_{\mathrm{end}}-t_0$} (12.3,2.4);
\draw[interval] (4.65,1.83) -- node[above,inner sep=2pt] {TPOT $=(t_{\mathrm{end}}-t_{\mathrm{obs}})/(D-1)$} (12.3,1.83);
\node[above right,inner sep=3pt] at (1.5,1.35) {$t_0$};
\node[below right,inner sep=3pt] at (4.65,1.35) {$t_{\mathrm{obs}}$};
\node[above right,inner sep=3pt] at (12.3,1.35) {$t_{\mathrm{end}}$};
\foreach \x in {1.5,4.65,12.3}
    \fill (\x,1.35) circle (1.4pt);
\draw[line width=0.5pt] (1.85,0.26) -- (1.85,0.85);
\draw[line width=0.5pt] (4.05,0.26) -- (4.05,0.85);
\draw[interval] (1.85,0.75) -- node[above,inner sep=2pt] {TTFT} (4.05,0.75);
\end{tikzpicture}
}
\caption{Inference timing for $D>1$ output tokens. $t_{\mathrm{obs}}$ marks CPU observation of the first token.}
\label{fig:inference-timeline}
\end{figure}
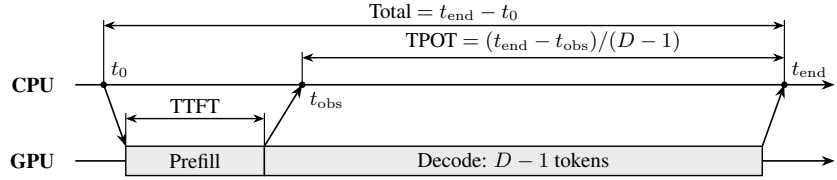

\paragraph{End-to-end timing.}
Figure~\ref{fig:inference-timeline} defines the timing intervals for fixed-length greedy generation without early stopping.
Argmax and token feedback stay on the main CUDA stream, with no per-token host waits.
TTFT uses GPU events; total latency and TPOT use the CPU wall clock, so TPOT includes observer scheduling effects and final delivery rather than measuring isolated decode kernels.
Each latency metric reports its own median over ten runs; speedup is the ratio of FP16's median total latency to the method's.
Output throughput is total generated tokens divided by total wall time across runs, including prompt processing.

\paragraph{Memory accounting.}
All plots reconstruct quantized weights on demand, without retaining additional full-precision copies between requests; online activation transforms remain timed.
For FP16, AWQ, SPHQuant, and SAGE, the weight-retention switch has no effect; the figures consistently use runs with that switch disabled.
Peak memory is \texttt{torch.cuda.max\_memory\_allocated}, reset before model initialization and read after measurement, so compilation and warmup peaks are included.
It excludes CUDA-context and other allocations outside PyTorch's allocator: exceeding a device's capacity precludes deployment, but falling below it does not guarantee a fit.
Main-text peaks are maxima over the two main workloads; Fig.~\ref{fig:inference-efficiency}(d) instead reports packed model storage, including unquantized parameters.

\paragraph{Retaining reconstructed weights.}
QuIP\#, QTIP, and UniSVQ reconstruct each layer's weights before prompt processing.
Discarding each reconstructed matrix after use saves memory; retaining all such matrices across requests avoids repeated reconstruction but adds a 16-bit copy of the quantized layers alongside the compressed weights.
With retention, first-token latencies for P128/D1024 and P1024/D128 are QuIP\# 21.7/87.3\,ms, QTIP 21.9/89.7\,ms, and UniSVQ 18.9/68.3\,ms, respectively.
Peak allocations rise to 8.80--8.86\,GiB.

\begin{table}[ht]
\centering
\caption{Short non-streaming workload: 128 input and 128 output tokens, without retained weight copies. First token is GPU TTFT; next token is CPU-observation-based TPOT, each reported as a median over runs.
SAGE achieves a $2.63\times$ end-to-end speedup over FP16.}
\label{tab:inference-short}
\footnotesize
\setlength{\tabcolsep}{4pt}
\begin{tabular}{cccccc}
\toprule
Method & First token (ms) & Next token (ms) & Total (s) & Output (tokens/s) & Peak (GiB) \\
\midrule
FP16 & 14.28 & 9.869 & 1.268 & 101.0 & 7.665 \\
AWQ & 13.18 & 4.115 & 0.536 & 238.7 & 2.611 \\
QuIP\# & 51.96 & 5.719 & 0.779 & 164.4 & 1.714 \\
QTIP & 48.35 & 5.638 & 0.765 & 167.3 & 1.898 \\
UniSVQ & 21.53 & 4.159 & 0.550 & 232.7 & 1.714 \\
SPHQ & 125.13 & 3.699 & 0.595 & 215.1 & 1.921 \\
SAGE & 17.14 & 3.661 & 0.483 & 265.2 & 1.857 \\
\bottomrule
\end{tabular}
\end{table}

\subsection{End-to-End Numerical Consistency}
\label{sec:inference-consistency}
Table~\ref{tab:inference-consistency} compares each quantized method with its own dense quantize--dequantize (QDQ) reference, obtained by independently decoding its packed weights.
The 4-token references use eager execution; the 32- and 128-token references use compiled execution.
Decode relative L2 is the maximum over the three steps following prefill; maximum absolute error covers all four steps.
The Dense row instead compares compiled and eager execution with the original pretrained weights.

\begin{table}[ht]
\centering
\caption{End-to-end numerical consistency. Lower relative L2 is better.}
\label{tab:inference-consistency}
\footnotesize
\setlength{\tabcolsep}{6pt}
\begin{tabular}{ccccc}
\toprule
Method & \shortstack{Prefill\\tokens} & \shortstack{Prefill\\relative L2 (\%)} & \shortstack{Decode maximum\\relative L2 (\%)} & \shortstack{Maximum\\absolute error} \\
\midrule
Dense & 4 & 0.1192 & 0.2266 & 0.01660 \\
SPHQuant & 4 & 0.8494 & 1.0412 & 0.16016 \\
QuIP\# & 4 & 0.8565 & 1.2647 & 0.21484 \\
QTIP & 4 & 1.0206 & 1.3674 & 0.19336 \\
UniSVQ & 4 & 5.7379 & 7.5981 & 1.02002 \\
SAGE & 4 & \textbf{0.6233} & \textbf{0.7746} & \textbf{0.11914} \\
\midrule
SAGE & 32 & 0.8459 & 0.8986 & 0.14453 \\
\midrule
QuIP\# & 128 & 4.4006 & 6.2508 & 0.89844 \\
QTIP & 128 & 3.0801 & 3.5425 & 0.50781 \\
UniSVQ & 128 & 20.0390 & 20.1103 & 3.22803 \\
SAGE & 128 & \textbf{1.0376} & \textbf{1.0906} & \textbf{0.16406} \\
\bottomrule
\end{tabular}
\end{table}

\subsection{Experimental Setting Details}
\label{sec:evaluation-settings}

\paragraph{Serving.}
For the accuracy results in Table~\ref{tab:main-results}, we serve each model with vLLM 0.19.0~\citep{vllm} on a single RTX Pro 6000, with prefix caching, chunked prefill, and asynchronous scheduling enabled.
Context limits are 262{,}144 tokens for Qwen3.5-4B, 65{,}536 for Qwen3-30B-A3B, and 131{,}072 for the remaining models.
The Ministral models use the Transformers backend and Mistral tokenizer.
Automatic tool calling uses the \texttt{hermes} parser for Qwen3, \texttt{qwen3\_coder} for Qwen3.5, and \texttt{mistral} for Ministral; Qwen3.5 additionally uses the \texttt{qwen3} reasoning parser.

\paragraph{Evaluation.}
We use EvalScope 1.10.0 through the OpenAI-compatible API, following each model's official model card for sampling hyperparameters.
Qwen3.5 runs with thinking enabled and a maximum of 16{,}384 generated tokens, including reasoning; all other models allow 8{,}192 generated tokens.
Evaluation concurrency is 16 for Qwen3-30B-A3B and 32 otherwise.
We fix the generation and dataset-shuffling seeds to 42 and evaluate BFCL-v3 in function-calling mode.
Code execution uses network-isolated Docker sandboxes, each limited to one CPU and 512\,MiB of memory.

\subsection{Calibration Data Construction}
\label{appendix_calibration_data}

\paragraph{Prompt Collection and Filtering.}
Our calibration prompts are drawn from publicly available datasets, including OpenCodeInstruct~\citep{opencodeinstruct}, the training splits of GSM8K~\citep{gsm8k} and MATH~\citep{math}, DeepMath-103K~\citep{deepmath}, OpenHermes-2.5\footnote{\url{https://huggingface.co/datasets/teknium/OpenHermes-2.5}}, UltraChat~\citep{ultrachat}, ShareGPT\footnote{\url{https://huggingface.co/datasets/Aeala/ShareGPT_Vicuna_unfiltered}}, and When2Call~\citep{ross2025when2call}.
The collection covers code, mathematics, reasoning, instruction following, tool use, and general conversation.
We rigorously screen calibration prompts against all evaluation sets and remove any detected overlaps.
Before response generation, we also exclude prompts that cannot be rendered by the target model's chat template or leave insufficient context for response generation.
We sample the six domains with a target prompt ratio of $13:13:13:10:9:6$ using a fixed random seed and interleave the sampled prompts across domains.
These proportions apply to prompt counts rather than token counts.

\paragraph{Model-Specific Response Generation.}
We retain the source prompts and regenerate responses using each target model.
For each model being quantized, we apply its own chat template and generate responses with its full-precision checkpoint using greedy decoding.
For thinking models, we enable thinking mode and retain the generated reasoning traces alongside the final answers.
We preserve the template's role markers and concatenate each prompt with its response.
These sequences provide inputs for activation statistics collection and blockwise reconstruction without requiring ground-truth answer labels.

\paragraph{Sequence Packing.}
We concatenate the tokenized conversations, inserting an end-of-sequence token where needed, and partition the resulting streams into contiguous blocks of 2048 tokens.
A block may contain multiple conversations or a segment of a longer conversation.
Training and held-out streams are packed separately, yielding 1920 training blocks and 128 validation blocks.
This fixed token budget is used across models, while the tokenizer, chat template, and generated responses remain specific to each target model.

\subsection{Spherical Vector Quantization Preliminaries}
\label{sec:prelim_appendix}

We supplement Sec.~\ref{sec:prelim} with the reconstruction derivation, continuous parameterization, radius-grid construction, and storage accounting.

\paragraph{Direction and Radius Reconstruction.}
Since signs are preserved, the reconstruction error in Eq.~(\ref{eq:reconstruction}) reduces to approximating $\mathbf{p}=|\mathbf{w}|$ by a scaled nonnegative codeword.
For a unit codeword $\mathbf{c}_k$, the error decomposes as:
\begin{equation}
    \|\mathbf{p}-\hat{r}\mathbf{c}_k\|_2^2
    =
    \|\mathbf{p}\|_2^2
    -(\mathbf{p}^{\top}\mathbf{c}_k)^2
    +(\hat{r}-\mathbf{p}^{\top}\mathbf{c}_k)^2.
    \label{eq:appendix-error-decomposition}
\end{equation}
The first two terms measure the residual with an unrestricted continuous radius, whose optimum is $\tilde{r}_k^\star=\mathbf{p}^{\top}\mathbf{c}_k$.
Because both $\mathbf{p}$ and $\mathbf{c}_k$ are nonnegative, selecting the smallest residual is equivalent to maximizing cosine similarity.
The last term captures the additional error when the fitted radius is quantized.
Thus, direction assignment with a continuous radius and selection after radius quantization are distinct problems.
The activation-weighted extension yields Eq.~(\ref{eq:weighted-assign}).

\paragraph{Continuous Parameterization.}
Following SPHQuant~\citep{sphquant}, each codeword is parameterized by $v-1$ hyperspherical angles:
\begin{equation}
    c_1=\cos\theta_1,\qquad
    c_j=\left(\prod_{i=1}^{j-1}\sin\theta_i\right)\cos\theta_j
    \quad (2\leq j<v),\qquad
    c_v=\prod_{i=1}^{v-1}\sin\theta_i.
    \label{eq:appendix-angles}
\end{equation}
Unconstrained logits $\alpha_i$ determine the angles through:
\begin{equation}
    \theta_i
    =
    \epsilon+\left(\frac{\pi}{2}-2\epsilon\right)
    \operatorname{sigmoid}(\alpha_i),
    \label{eq:appendix-angle-logits}
\end{equation}
where $\epsilon=10^{-4}$ keeps angles away from the boundaries.
This mapping produces nonnegative unit codewords by construction.
Continuous radii are parameterized as $\tilde{r}=\operatorname{softplus}(\rho)$, ensuring positivity.
Blockwise optimization updates $\rho$ and the codebook logits while retaining the discrete assignments.

\paragraph{Shared Radius Grid.}
Each group contains $G/v=16$ consecutive subvectors within a weight row.
The grid offset and step in Eq.~(\ref{eq:radiusgrid}) are computed from their continuous fitted radii:
\begin{equation}
    \mu_g=\min_{i\in g}\tilde{r}_i,
    \qquad
    \delta_g=
    \frac{\max_{i\in g}\tilde{r}_i-\mu_g}{2^V-1}.
    \label{eq:appendix-grid-parameters}
\end{equation}
These parameters are recomputed from the current radii during optimization.
If all radii in a group coincide, the reconstruction is their common value.
The forward pass uses quantized radii, while a straight-through estimator passes gradients to the continuous radii.
Because the grid is shared, its resolution depends on the fitted radius range of the entire group.

\paragraph{Encoding and Storage Costs.}
The 2.25 BPW reported in the main text refers to the sign, direction-index, and radius-code payload in Eq.~(\ref{eq:budget}).
For a matrix with $N=d_{\mathrm{out}}d_{\mathrm{in}}$ weights, storing each grid parameter in $b_g$ bits and each codeword component in $b_c$ bits gives:
\begin{equation}
    \mathrm{BPW}_{\mathrm{matrix}}
    =
    \frac{B}{v}
    +\frac{2b_g}{G}
    +\frac{Kvb_c}{N},
    \label{eq:appendix-storage}
\end{equation}
excluding permutation and format metadata.
FP16 grid parameters contribute an additional $0.25$ BPW at $G=128$, bringing the subtotal to $2.50$ BPW before codebook and other metadata.
The codebook overhead depends on both $K$ and matrix size.
Accordingly, ABS preserves the per-subvector encoding budget, while total storage also reflects the selected codebooks and deployment metadata.

\subsection{SAGE Quantization Algorithm}
\label{sec:sage-algorithm}

\noindent\begin{minipage}{\linewidth}
\small
\hrule
\vspace{4pt}
\textbf{Algorithm 1: SAGE quantization}
\vspace{4pt}
\hrule
\begin{tabbing}
\hspace{1em}\=\hspace{1em}\=\hspace{1em}\=\kill
\textbf{Input:} Full-precision model and non-evaluation calibration prompts. \\
Pack prompts and greedy full-precision responses into calibration sequences $\mathcal{D}_{\mathrm{cal}}$. \\
Run full-precision calibration; estimate each matrix's input-channel energies $e_c:=\mathbb{E}[x_c^2]$. \\
\textbf{for each} weight matrix $W$ to quantize \textbf{do} \\
\> Sort channels by decreasing $e_c$; interleave eight equal-sized tiers to obtain $\pi$ (ESG). \\
\> Permute columns by $\pi$; split rows into 8-vectors, storing signs $s$ and magnitudes $p$. \\
\> Set position weights $\omega_i\propto e_{a_i}^{0.75}$, normalized to unit mean within each subvector. \\
\> \textbf{for each} $(K,V)\in\{(64,4),(128,3),(256,2)\}$ \textbf{do} \\
\>\> Fit $K$ nonnegative unit codewords by activation-weighted spherical $k$-means. \\
\>\> For each $p$, choose $z,\tilde r$ minimizing $\sum_i\omega_i(p_i-\tilde r c_{z,i})^2$. \\
\>\> Round radii to $2^V$ uniform levels spanning their range in each 128-weight group. \\
\>\> Score $\hat w:=\hat r(s\odot c_z)$ in original channel order by $\sum_c e_c\|W_{:,c}-\widehat W_{:,c}\|_2^2$. \\
\> \textbf{end for} \\
\> Retain the lowest-score candidate to initialize the quantized matrix (ABS). \\
\textbf{end for} \\
Initialize $x_1^{\mathrm{q}}:=x_1^{\mathrm{fp}}$ from the calibration embeddings. \\
\textbf{for} $\ell=1,\ldots,L$ \textbf{do} \\
\> Cache full-precision targets $F_\ell(x_\ell^{\mathrm{fp}})$. \\
\> Fix signs, direction indices, permutations, and bit splits in the quantized block $\widehat F_\ell$. \\
\> \textbf{for each} minibatch over all optimization epochs \textbf{do} \\
\>\> Recompute radius grids and reconstruct quantized weights. \\
\>\> Use Adam to minimize mean squared error between $\widehat F_\ell(x_\ell^{\mathrm{q}})$ and cached targets. \\
\>\> Update only radii and codebook angles; pass gradients through radius rounding. \\
\> \textbf{end for} \\
\> Propagate $x_{\ell+1}^{\mathrm{q}}:=\widehat F_\ell(x_\ell^{\mathrm{q}})$ and $x_{\ell+1}^{\mathrm{fp}}:=F_\ell(x_\ell^{\mathrm{fp}})$. \\
\textbf{end for} \\
\textbf{Output:} Signs, direction and radius codes, grids, codebooks, and permutations; \\
\> retain unquantized model parameters unchanged.
\end{tabbing}
\hrule
\end{minipage}

\subsection{Dialog Examples}

\begingroup
\fontencoding{T1}\selectfont
\newcommand{\DialogModel}[1]{\textbf{\raisebox{-0.15\height}{\includegraphics[height=1.875ex]{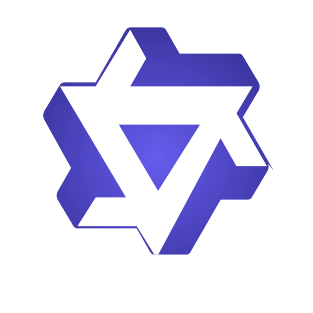}}\ #1:}\ }
\newcommand{\DialogReason}[1]{{\color{black!65}#1}\ }
\newcommand{\DialogOutput}[1]{#1}
\newcommand{\DialogHighlight}[1]{\begingroup\setlength{\fboxsep}{0pt}\colorbox{purple!9}{\rule[-0.3\baselineskip]{0pt}{\baselineskip}\kern2pt\textcolor{purple!80!black}{\fontencoding{T1}\rmfamily\selectfont#1}\kern2pt}\endgroup}
\newcommand{\DialogAnswer}[1]{\DialogHighlight{\{\textquotedbl answer\textquotedbl: \textquotedbl#1\textquotedbl\}}}
\newcommand{\DialogMissing}[1]{\textcolor{black!60}{\emph{#1}}}
\newcommand{\DialogSeparator}{\noalign{\vskip 1pt}\hdashline\noalign{\vskip 1pt}}
\normalsize
\setlength{\tabcolsep}{7pt}
\setlength{\dashlinedash}{2pt}
\setlength{\dashlinegap}{2pt}
\setlength{\LTleft}{0pt}
\setlength{\LTright}{0pt}
\setlength{\LTcapwidth}{\textwidth}
\renewcommand{\arraystretch}{0.7}
\begin{longtable}{>{\fontsize{7}{7}\selectfont}p{\dimexpr\textwidth-2\tabcolsep\relax}}
\caption{Qwen3.5-4B dialogues with native BF16 or quantized--dequantized (QDQ) weights. Both prompts use seed 42, temperature 1.0, and thinking mode. Generation limits, including reasoning, are 512 tokens for Prompt 1 and 256 for Prompt 2. Gray text denotes reasoning.}\label{tab:dialog-examples} \\
\toprule
\endfirsthead
\multicolumn{1}{@{}l@{}}{\small Table~\thetable\ (continued)} \\
\toprule
\endhead
\midrule
\multicolumn{1}{@{}r@{}}{\scriptsize Continued on next page} \\
\endfoot
\bottomrule
\endlastfoot
\rowcolor{black!7}
{\textbf{\raisebox{-0.15\height}{\includegraphics[height=2.25ex]{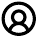}}\ Prompt 1} Five numbers have an average of 18. If the number 22 is removed, what is the average of the remaining four numbers? Choices: A. 16 B. 17 C. 18 D. 19. Please show your choice in the \texttt{answer} field with only the choice letter, e.g., \{\textquotedbl answer\textquotedbl: \textquotedbl C\textquotedbl\}.} \\*
\midrule
\DialogModel{BF16}
\DialogReason{The user wants me to solve a math problem and provide the answer in a specific format. \textbf{1. Understand the Problem:} There are 5 numbers. The average (arithmetic mean) of these 5 numbers is 18. One number, which is 22, is removed from the set. I need to find the new average of the remaining 4 numbers. \textbf{2. Formulate Equations:} Let the five numbers be $x_1, x_2, x_3, x_4, x_5$. The sum of the five numbers is $S_5$. Average formula: $\text{Average} = \frac{\text{Sum}}{\text{Count}}$\quad Given: $\frac{S_5}{5} = 18$. Therefore, $S_5 = 18 \times 5$. Calculate $S_5$: $18 \times 5 = 90$. So, the sum of the original 5 numbers is 90. Now, one number is removed. This number is $x_k = 22$. The sum of the remaining 4 numbers is $S_4 = S_5 - x_k$. Calculate $S_4$: $S_4 = 90 - 22 = 68$. The count of the remaining numbers is 4. The new average is $\text{Average}_{new} = \frac{S_4}{4}$. Calculate $\text{Average}_{new}$: $\frac{68}{4}$. $68 / 4 = 17$. \textbf{3. Check the Choices:} A. 16\quad B. 17\quad C. 18\quad D. 19\quad My calculated answer is 17, which corresponds to Choice B. \textbf{4. Structure the Output:} State the steps clearly. Conclude with the final choice. Format as requested: \{"answer": "B"\}. \textbf{Double Check:} Original Sum = $5 \times 18 = 90$. Remove 22}
\DialogMissing{(TRUNCATED)} \\
\DialogSeparator
\DialogModel{QuIP\#}
\DialogReason{The user wants me to solve a math problem and provide the answer in a specific format. 1. \textbf{Understand the problem}: There are five numbers. The average of these five numbers is 18. One number, specifically 22, is removed. The remaining four numbers must have an average value. I need to determine what that average is from the given choices (A=16, B=17, C=18, D=19). 2. \textbf{Analyze the data}: Let $n$ be the number of items in a set. Let $x_1, x_2, ..., x_n$ be the elements. The formula for the average is: $\text{average} = \frac{\text{sum}}{n}$. Here, we have 5 numbers whose sum equals $5 \times 18 = 90$. When 22 is removed, there are 4 numbers left. The new sum is 90 - 22 = 68. The new average would be $68 / 4 = 17$. 3. \textbf{Wait, let me check my reasoning.} If the original average is 18, with 5 numbers, their total sum is 90. If one number, 22, is removed, the remaining 4 numbers sum to $90 - 22 = 68$. Average of the remaining 4 numbers = $68/4 = 17$. This seems like a straightforward calculation based on averages. However, does this logic work? The problem doesn't say ``one of the numbers was 22''. It says ``the number 22 is removed''. Since there were 5 numbers with an average of 18, the sum is indeed 90. Is it possible for the average of the remaining 4 numbers to still be 18? The question asks which choice is correct. If the sum is 90, removing 22 leaves 68. 68 divided by 4 is 17. So the answer should be 17.\quad 4}
\DialogMissing{(TRUNCATED)} \\
\DialogSeparator
\DialogModel{QTIP}
\DialogReason{Thinking Process: 1. \textbf{Calculate total sum of the initial five numbers:} Since the average is 18 and there are 5 numbers, the total sum is $18 \times 5 = 90$. 2. \textbf{Remove the number 22 from the set.} The remaining four numbers have a sum of $90 - 22 = 68$. 3. \textbf{Calculate the new average:} Divide the remaining sum by the count of remaining numbers (4). New average = $68 / 4 = 17$. 4. \textbf{Match with choices:} Option B corresponds to 17. 5. \textbf{Format output:} Provide the choice letter in the answer field as requested. \textbf{Step-by-Step Calculation:} Sum of original numbers = Average $\times$ Count = $18 \times 5 = 90$. After removing 22, Sum = $90 - 22 = 68$. Remaining count = 4. New Average = $68 / 4 = 17$. Choice B matches 17. Answer should be ``B''.}
\DialogOutput{\DialogAnswer{B}} \\
\DialogSeparator
\DialogModel{UniSVQ}
\DialogReason{The user wants me to know the average of the remaining 4 numbers after removing the number 22. The five original numbers were used to get an average of 18. What are the implications? A. It's about a math problem where I can't remember what it was when I was at school or work? B. It's about statistics: What is the mean of a series? C. It's a question that needs to be answered without knowing the answer. D. It could be related to the average value being something else entirely. E. The question might have nothing to do with statistics, other than the fact that the average is calculated differently in some contexts.}
\DialogMissing{(NO FINAL)} \\
\DialogSeparator
\DialogModel{SPHQuant}
\DialogReason{The user wants me to solve a math problem about averages. 1. \textbf{Analyze the given information:} There are five numbers. Their average is 18. One of the numbers (22) is removed. I need to find the average of the remaining four numbers. 2. \textbf{Formulate the relationship between average and sum:} Average = Sum / Count\quad Sum = Average * Count\quad 3. \textbf{Calculate the total sum of the original five numbers:} Count = 5\quad Average = 18\quad Sum = 5 * 18\quad 4. \textbf{Perform the multiplication:} $5 \times 18 = 90$\quad 5. \textbf{Determine the sum of the remaining four numbers:} Original Sum - Removed Number\quad Sum = 90 - 22\quad $90 - 22 = 68$\quad 6. \textbf{Calculate the new average:} New Average = Sum of remaining numbers / Count of remaining numbers\quad New Count = 4\quad New Average = 68 / 4\quad 7. \textbf{Perform the division:} $68 / 4 = 17$\quad 8. \textbf{Verify with the choices:} A. 16\quad B. 17\quad C. 18\quad D. 19\quad The calculated result matches choice B. 9. \textbf{Format the output:} The user requested the answer in the answer field with only the choice letter. Format: \{"answer": "B"\}\quad 10. \textbf{Final check:} Original sum: $5 \times 18 = 90$. Remove 22: $90 - 22 = 68$. New count: $5 - 1 = 4$. New average: $68 / 4 = 17$. Choice is B. Constraint: ``Please show your choice in the answer field with only the choice}
\DialogMissing{(TRUNCATED)} \\
\DialogSeparator
\rowcolor{purple!4}
\DialogModel{SAGE}
\DialogReason{The user wants me to solve a math problem and provide the answer in a specific format. \textbf{Problem:} Five numbers have an average of 18. One number (22) is removed. What is the average of the remaining four numbers? \textbf{Step 1: Calculate the sum of the original five numbers.} Average = Sum / Count\quad $18 = \text{Sum} / 5$\quad $\text{Sum} = 18 \times 5$\quad $\text{Sum} = 90$\quad \textbf{Step 2: Calculate the sum of the remaining four numbers.} The number 22 is removed. New Sum = Original Sum - 22\quad New Sum = $90 - 22$\quad New Sum = 68\quad \textbf{Step 3: Calculate the new average.} New Average = New Sum / New Count\quad New Average = $68 / 4$\quad New Average = 17\quad \textbf{Step 4: Format the output.} The user requested the choice letter in the answer field. Choices: A. 16, B. 17, C. 18, D. 19. My calculated answer is 17, which corresponds to Choice B. The format requires \{"answer": "B"\}. \textbf{Double Check:} Original sum = 90. Remove 22 -\textgreater{} Remainder sum = 68. Count = 4. Average = 68 / 4 = 17. Matches Choice B. \textbf{Final Output Generation:} Construct the JSON response.}
\DialogOutput{The sum of the five numbers is $18 \times 5 = 90$. After removing 22, the new sum is $90 - 22 = 68$. The average of the remaining four numbers is $68 / 4 = 17$. This corresponds to Choice B. \DialogAnswer{B}} \\
\midrule
\rowcolor{black!7}
{\textbf{\raisebox{-0.15\height}{\includegraphics[height=2.25ex]{figs/user-icon.pdf}}\ Prompt 2} Do not enumerate steps or analyze options. Answer immediately. Which day follows Tuesday? Output the lowercase name of the day, followed by an exclamation mark, with no other text.} \\*
\midrule
\DialogModel{BF16}
\DialogReason{Thinking Process: 1. \textbf{Analyze the Request:} Task: Identify the day that follows Tuesday. Constraint 1: Do not enumerate steps. Constraint 2: Do not analyze options. Constraint 3: Answer immediately. Format: Lowercase name of the day, followed by an exclamation mark, with no other text. 2. \textbf{Determine the Answer:} Day sequence: Monday -\textgreater{} Tuesday -\textgreater{} Wednesday... Day following Tuesday is Wednesday. 3. \textbf{Format the Output:} Name: wednesday\quad Punctuation: !\quad Combined: wednesday!\quad Check constraints: No extra text, lowercase, exclamation mark at the end. 4. \textbf{Final Verification:} Is it lowercase? Yes (wednesday). Is there an exclamation mark? Yes (!). Is there any other text? No. Did I enumerate or analyze? No. 5. \textbf{Construct Output:} wednesday! 6. \textbf{Final Decision:} Output \textquotedbl wed}
\DialogMissing{(TRUNCATED)} \\
\DialogSeparator
\DialogModel{QuIP\#}
\DialogOutput{\DialogHighlight{Wednesday!}} \\
\DialogSeparator
\DialogModel{QTIP}
\DialogReason{Yes!}
\DialogOutput{\DialogHighlight{yes!}} \\
\DialogSeparator
\DialogModel{UniSVQ}
\DialogReason{The user wants me to know what day follows Tuesday. The answer is Wednesday. It's a straightforward question that requires an immediate response without any further analysis or steps.}
\DialogMissing{(NO FINAL)} \\
\DialogSeparator
\DialogModel{SPHQuant}
\DialogReason{Thinking Process: 1. \textbf{Analyze the Request:} Task: Identify the day that follows Tuesday. Constraint 1: Do not enumerate steps or analyze options. (No list, no explanation). Constraint 2: Answer immediately. Constraint 3: Output the lowercase name of the day. Constraint 4: Follow with an exclamation mark (!). Constraint 5: No other text. 2. \textbf{Determine the Day:} Current day is Tuesday. The day following Tuesday is Wednesday. 3. \textbf{Format the Output:} Lowercase name: \textquotedbl Wednesday\textquotedbl\quad Exclamation mark: \textquotedbl!\textquotedbl\quad Combined: \textquotedbl Wednesday!\textquotedbl\quad No extra text. 4. \textbf{Review Constraints:} Did I enumerate? No. Did I analyze options? No. Is it immediate? Yes. Is it lowercase? Yes (\textquotedbl Wednesday\textquotedbl). Is there an exclamation mark? Yes (\textquotedbl!\textquotedbl). Is there any other text}
\DialogMissing{(TRUNCATED)} \\
\DialogSeparator
\rowcolor{purple!4}
\DialogModel{SAGE}
\DialogReason{Thinking Process: 1. \textbf{Analyze the Request:} Task: Determine which day follows Tuesday. Constraint 1: Do not enumerate steps or analyze options. Constraint 2: Answer immediately. Constraint 3: Output the lowercase name of the day, followed by an exclamation mark. Constraint 4: No other text. 2. \textbf{Determine the Answer:} The day following Tuesday is Wednesday. Format required: lowercase \textquotedbl wednesday!\textquotedbl\quad 3. \textbf{Check Constraints:} Is it a single output? Yes. Is it lowercase? Yes (\textquotedbl wednesday\textquotedbl). Is there an exclamation mark? Yes (\textquotedbl!\textquotedbl). Is there any analysis or enumeration? No. 4. \textbf{Final Output Formulation:} wednesday! 5. \textbf{Final Review:} Does this meet all criteria? Yes.cw}
\DialogOutput{\DialogHighlight{wednesday!}} \\
\end{longtable}
\endgroup



\end{document}